\documentclass{article}
\usepackage[preprint, nonatbib]{nips_2018}
\usepackage[numbers]{natbib}

\usepackage[utf8]{inputenc} 
\usepackage[T1]{fontenc}    
\usepackage{hyperref}       
\usepackage{url}            
\usepackage{booktabs}       
\usepackage{amsfonts}       
\usepackage{nicefrac}       
\usepackage{microtype}      
\usepackage{amsmath}
\usepackage{amssymb}
\usepackage{bm}
\usepackage{xcolor}
\usepackage{graphicx}
\usepackage{subfig}
\usepackage{comment}
\usepackage{algorithm,algcompatible,lipsum}
\usepackage{multirow}
\usepackage{wrapfig}
\usepackage{float}
\usepackage{multicol}

\newcommand{\real}{\mathbb{R}}
\newcommand{\p}{\text{p}}
\newcommand{\E}{\textbf{\text{E}}}
\newcommand{\Var}{\textbf{\text{Var}}}
\newcommand{\Id}{\textbf{\text{I}}}
\newcommand{\Gaussian}[2]{\mathcal{N}(#1, #2)}
\newcommand{\de}{\text{d}}

\algnewcommand{\lst}{\texttt{lst}}
\algnewcommand{\slst}{\texttt{slst}}
\algnewcommand{\SEND}{\textbf{send}}

\DeclareMathOperator{\erf}{erf}

\title{Propagating Uncertainty through the tanh Function with Application to Reservoir Computing}

\author{
  Manan Gandhi \\
  Georgia Institute of Technology\\
  \texttt{mgandhi@gatech.edu} \\
  \And
  Keuntaek Lee \\
  Georgia Institute of Technology \\
  \texttt{keuntaek.lee@gatech.edu} \\
  \AND
  Yunpeng Pan \\
  JD.COM American Technologies Corporation\\
  \texttt{yunpeng.pan@jd.com} \\
  \And
  Evangelos A. Theodorou \\
  Georgia Institute of Technology \\
  \texttt{evangelos.theodorou@gatech.edu} \\
}

\begin{document}


\maketitle

\begin{abstract}
Many neural networks use the $\tanh$ activation function, however when given a probability distribution as input, the problem of computing the output distribution in neural networks with $\tanh$ activation has not yet been addressed. One important example is the initialization of the echo state network in reservoir computing, where random initialization of the reservoir requires time to wash out the initial conditions, thereby wasting precious data and computational resources. Motivated by this problem, we propose a novel solution utilizing a moment based approach to propagate uncertainty through an Echo State Network to reduce the washout time. In this work, we contribute two new methods to propagate uncertainty through the $\tanh$ activation function and propose the Probabilistic Echo State Network (PESN), a method that is shown to have better average performance than deterministic Echo State Networks given the random initialization of reservoir states. Additionally we test single and multi-step uncertainty propagation of our method on two regression tasks and show that we are able to recover similar means and variances as computed by Monte-Carlo simulations.
\end{abstract}
\section{Introduction}
Neural networks can often have inputs that are uncertain. Input uncertainty can come from measurement error, adversarial noise \citep{szegedy2013intriguing}, or even from output feedback. The majority of work on the subject of uncertainty in neural networks revolves around uncertainty in the model, not necessarily uncertainty in the input itself. Bayesian neural networks perform inference using a prior over the weights of the network, while dropout \citep{Gal2016DropoutRNN} samples iterations of the network with a probabilistic mask to build the posterior distribution of the model. In both cases the output uncertainty emerges as an explicit function of the model uncertainty, while the uncertainty of the input is classified as aleatoric and not explicity propagated through the model. For non-parametric probabilistic methods, such as Gaussian process regression \citep{rasmussen2004gaussian}, special care must be taken for multi-step prediction with uncertain inputs \citep{girard2003gaussian}. In this paper, we focus on addressing the problem of uncertainty propagation through the $\tanh$ function, which is a popular choice for activation function in neural networks, in particular the Echo State Network.

In light of the challenges in input uncertainty propagation and their role in recurrent neural networks, we aim to contribute the following:
\begin{itemize}
\item A theoretical and numerical analysis for 3 methods of propagating input uncertainty through the $\tanh$ activation function, with an extension to other nonlinear activation functions. 
\item A new method, named the Probabilistic Echo State Network (PESN), which aims to reduce the time required to achieve the echo state property.
\end{itemize}

\textbf{Related Works:} There exists recent work on propagating input uncertainty through feed-forward bayesian neural networks \citenum{shekhovtsov2018feed}. Here, the authors perform approximate inference to propagate uncertain inputs through feedforward neural networks for classification tasks. We differentiate our work in three fundamental ways: 1) we consider the tanh function which is not addressed in \citep{shekhovtsov2018feed}, 2) a novel contribution utilizing splines to propagate gaussian input uncertainty through continuous activations and 3) a general focus on improving the reservoir computing framework. To the best of the authors' knowledge we are the first to utilize a spline approximation to the integrand of an expectation of a gaussian in order to perform approximate inference. Similar ideas include \citep{langrock2015nonparametric}, where the authors utilize b-splines to approximate the density in order to perform approximate inference. Work in improving reservoir computing has traditionally focused on the structure of the reservoir, such as optimizing hyperparameters \citep{jaeger2007optimization} or finding the minimum reservoir size \citep{rodan2011minimum}. This work is one of the first to attempt to reduce the time required to converge to the echo state property.
\subsection{Reservoir Computing}
Reservoir computing (RC) is a paradigm for training recurrent neural networks (RNNs) \citep{lukovsevivcius2009reservoir}. It was  introduced in early 2000's by Jaeger under the name `Echo State Networks' (ESN) for time-series predictions \citep{jaeger2004harnessing}, and by Maass under the name `Liquid State Machines' \citep{maass2004computational} for modeling computation in biological networks of neurons. Despite being developed from very different communities, these two approaches are largely mathematically identical. In this work we focus on ESN since we aim to improve the efficiency of RNN learning for engineering applications.

ESN differs from other RNNs in terms of its training scheme. Generally, RC consists of two steps: 1) drive a network with sparse and fixed connections with an input and output sequence. 2) Train the output (or readout) layer so that the network output is similar to the teacher output. The readout layer is usually trained using regression techniques such as linear regression \citep{lukovsevivcius2009reservoir} and Gaussian process regression \citep{chatzis2011echo}. One problem is determining the initial state of the hidden layer (reservoir). According to the echo state property \citep{jaeger2001echo}, the effect of  initial conditions can be `washed out' therefore the state can be initialized randomly. However, there are three main drawbacks: first, a significant amount of training data is wasted because the initial period of training run needs to be discarded. Second, at test time,  an initial input sequence needs to be fed into the network before using it for prediction tasks. Third, the state forgetting property is usually not guaranteed so the performance of the ESN may still depend on the initial state.
 
\section{Propagating Uncertainty through the tanh}
In our work we analyze three distinct methods to propagate a gaussian input through the $\tanh$ activation function. The simplest and most well known method is simply Monte-Carlo (MC) sampling where we sample the input distribution then pass each sample through the activation to compute an estimate of the moments. While easy to implement and understand, Monte Carlo has an obvious drawback in terms of computational time. The second method utilizes a well known approximation to the $\tanh$ activation function in the form of the logistic cumulative distribution function. The connection between the logistic and gaussian distributions are utilized to approximate the mean of the activation output. The variance approximation fits the moments of the Gaussian pdf to the function $(1 - \tanh(x)^2)$ in order make the expectation tractable. The cross covariance terms are ignored in our derivation. While this method is certainly faster, the accuracy is limited by our approximations. The final method leverages spline approximations and analytical expressions for the expectation of polynomial functions. This method strikes a balance between computational complexity and accuracy by adjusting the width of the spline mesh. We compare absolute error of the moment approximations, computational complexity of the two analytical and spline methods, and provide error bounds for the spline approximation.

\subsection{Analytical Approximation to Mean and Variance}
First we relate the $\tanh$ function with the logistic cumulative distribution function CDF, and approximate the logistic distribution with an appropriate gaussian distribution. We assume that the dimensions of the logistically distributed random vector $x$ are independent given the location parameter $\bm{z}$. The expectation and variance of $x$ given $\bm{z}$ are $\E(x | \bm{z}) = \bm{z}, \Var(z | \bm{z}) = \frac{\pi^2}{12}\Id$ respectively. We can use moment matching to approximate $x$ with a gaussian 
\begin{eqnarray}
\p(x \geq 0 | \lambda, \xi) &=& \frac{1}{1 + \exp(-\frac{\lambda}{\xi})} \label{logistic cdf} \\
\implies \tanh(\bm{z}) &=& 2\big(\frac{1}{1 + \exp(-2\bm{z})}\big) - 1 \\
&=& 2 \p(x \geq 0 | \bm{z}, 0.5) - 1 \label{tanh relate}
\end{eqnarray}
Here the logistic distribution is parameterized by the location $\lambda$ and the scale $\xi$. The $\tanh$ function is a function of a logistic CDF with location $\lambda=\bm{z}$ and scale $\xi=0.5$. Additionally let us assume that $\bm{z} \sim \Gaussian{\bm{\mu}}{\bm{\sigma}}$. The mean of the distribution, $\p(\tanh(\bm{z}))$, is given by:
\begin{eqnarray}
\E(\tanh(\bm{z})) &\approx& \frac{2}{1 + \exp(-\frac{\bm{\mu_i}}{\sqrt{\frac{3}{\pi^2}\bm{\sigma_i} + \frac{1}{4}}})} - 1 = \mu_{\tanh(\bm{z}_i)} ~~,~~\text{for the  $i^{\text{th}}$ element of $\bm{z}$} \label{mean1}
\end{eqnarray}

Next, we ignore the cross covariance terms between the elementwise $\tanh$, and directly compute the diagonals of the output covariance matrix. The authors in \citep{bassett1998x2} investigated different approximations to the error function. One preliminary solution was the use of the tanh function to approximate the error function. We can take advantage of this approximation and the appropriate scaling factors to approximate the function $(1-\tanh(x))^2$. We use a Gaussian pdf where $\sigma^2 = \frac{2}{\pi}$ and $y=\frac{2}{\sqrt{\pi}}x$. Then we can again apply similar algebraic manipulations to compute the variance terms. The full derivation for the mean and variance expressions can be found in the supplementary material.
\begin{eqnarray}
\frac{2}{\sqrt{2\pi\sigma^2}}\exp(-\frac{y^2}{2\sigma^2}) &\approx& (1-\tanh(y^2)) \label{tanh2 approx}
\end{eqnarray}
\begin{eqnarray}
 \Var(\tanh \bm{z}_i) &\approx&  1 - \frac{2}{\sqrt{2\pi(\frac{2}{\pi} + \bm{\sigma}_{ii})}}\exp(-\frac{\bm{\mu}_i^2}{2(\frac{2}{\pi} + \bm{\sigma}_{ii})} ) - \label{variance1}  \mu_{\tanh(\bm{z}_i)}^2
\end{eqnarray}
%
%


%
\subsection{Spline Approximation to Mean and Variance}
In this section, we present a formulation for computing the moments of the $\tanh$ transformation utilizing a spline approximation of the argument of the expectation. Again let us assume that $\bm{z} \sim \Gaussian{\bm{\mu}}{\bm{\sigma}}$, the expectation and variance of the elementwise $\tanh$ are given by the following equations:
\begin{eqnarray}
\E( \tanh(\bm{z}_i)) &=& \int_{-\infty}^{\infty} \tanh(\bm{z}_i) \frac{1}{\sqrt{2 \pi \sigma_i}} \exp\big({\frac{-1}{2\sigma}(\bm{z}_i - \mu_i)^2}\big) \de \bm{z}_i \label{mean_tanh_spline}\\ \nonumber
\E \big(\tanh(\bm{z}_i) - \E (\tanh(\bm{z}_i))\big)^2 &=& 1 - \int_{-\infty}^{\infty} (1 - \tanh(\bm{z}_i)^2) \frac{1}{\sqrt{2 \pi \sigma_i}} \exp\big({\frac{-1}{2\sigma}(\bm{z}_i - \mu_i)^2}\big) \de \bm{z}_i - \\ && -\E (\tanh(\bm{z}_i))^2 \label{var_tanh_spline}
\end{eqnarray}
 Let us assume that we are dealing with scalar inputs to the activation function, since the vector case is handled elementwise and we are ignoring the off-diagonal elements of the input variance. Instead of direct numeric integration, we propose taking advantage of three facts. 
 \begin{itemize}
 \item The tails of the Gaussian pdf approach zero, thus when multiplied against a function with constant tails, such as the $\tanh$, the integrand also goes to zero. Thus we can approximate the integral over the real line using an integral over some compact subset centered around the mean of the input.
 \item The $\tanh$ function is continuous so we can uniformly approximate it with polynomials on the compact subset of integration.
 \item  We can derive analytic forms for the definite integral of a product of polynomials and exponentials.
 \end{itemize}
  As a result, we can derive expressions for the mean and variance in terms of these analytic integrals. Let $f(z)$ be a scalar valued function that is continuous on the real interval $[a, b]$. Additionally, let $P(z)$ be a piecewise continuous cubic polynomial that interpolates $f$ with $N$ nodes, where each node is the beginning of intervals $[z_j, z_{j+1}]$, where $z_0 = a$ and $z_N = b$.
\begin{eqnarray}
\E( f(z)) &\approx& \frac{1}{\sqrt{2 \pi \sigma}} \int_{a}^{b} P(z)  \exp\big({\frac{-1}{2\sigma}(z - \mu)^2}\big) \de z \\
&=& \frac{1}{\sqrt{2 \pi \sigma}} \sum_{j=0}^{N-1}  \int_{z_j}^{z_{j+1}} P(z) \exp\big({\frac{-1}{2\sigma}(z - \mu)^2}\big) \de z \\
&=& \frac{1}{\sqrt{2 \pi \sigma}} \sum_{j=0}^{N-1}  \int_{z_j}^{z_{j+1}} \sum_{k=0}^3 c_{jk} z^k  \exp\big({\frac{-1}{2\sigma}(z - \mu)^2}\big) \de z \\
&=& \frac{1}{\sqrt{2 \pi \sigma}} \sum_{j=0}^{N-1} \sum_{k=0}^3 c_{jk}  \int_{z_j}^{z_{j+1}} z^k    \exp\big({\frac{-1}{2\sigma}(z - \mu)^2}\big) \de z \label{spline}
\end{eqnarray}
Within the inner sum, the four terms can be computed analytically utilizing expressions for the integral of the product between polynomials and exponentials. We provide the full expressions for polynomials up to 3rd order in the supplementary material. The nodes and coefficients can be computed prior to evaluation of the network, and the integrals are computed once at every layer. This method is in fact general for any continuous function $f(z)$ whose tails are bounded, or polynomial outside of a compact set. Additionally, we can move forward to approximate higher moments of the output distribution and get expressions for the skew and kurtosis, at the cost of having spline approximations for $f(z)^3$ and $f(z)^4$. We provide expressions for higher moments and for alternative activation functions in the supplement. The mean and variance are given below, and Algorithm \ref{alg:spline} describes how to use this technique to compute the moments of the activation.
\begin{eqnarray}
\mu_{\tanh(\bm{\mu}_i)} &=& A_1 \label{mean} \\
\sigma_{\tanh(\bm{\mu}_i)} &=& A_2 - A_1^2 \label{variance} \\
A_1 &=& \int_{-\infty}^{\infty} f(\bm{z}_i) \frac{1}{\sqrt{2 \pi \bm{\sigma}_i}} \exp\big({\frac{-1}{2\bm{\sigma}_i}(\bm{z}_i - \bm{\mu}_i)^2}\big) \de \bm{z}_i \\
A_2 &=& \int_{-\infty}^{\infty} f(\bm{z}_i)^2 \frac{1}{\sqrt{2 \pi \bm{\sigma}_i}} \exp\big({\frac{-1}{2\bm{\sigma}_i}(\bm{z}_i - \bm{\mu}_i)^2}\big) \de \bm{z}_i
\end{eqnarray}
\subsection{Computational Complexity Analysis}
Next we look at the computational complexity of both the analytical and spline method in terms of the number of hidden states in the network. Let D be the number of hidden states in the network and let N be the number of mesh points for the spline approximation. Since the spline method makes frequent use of the error function, to approximate the operation count, we use the error function approximation introduced in \citep{Abramowitz1964erf}. We can see from Table \ref{table: Algorithm complexity analysis} that if we use the spline approximation of $\tanh$, the time complexity is $\mathcal{O}$(ND), while the analytical approximation of $\tanh$ has the complexity of $\mathcal{O}$(D). If we choose a fixed number of Monte Carlo samples $M$, then the Monte Carlo method is also linear with respect to reservoir $D$ since the $\tanh$ activation is element-wise and cross correlations are ignored. All 3 are linear in reservoir size and in practice we can see the this behavior of all 3 methods in Figure \ref{fig:time_complexity}.

\begin{table}[ht]
\begin{minipage}[b]{0.48\linewidth}
\centering
\begin{algorithm}[H]
    \caption{$(\bm{\mu}, \bm{\sigma})$ approximation of $\tanh(\bm{z})$ with spline method}
    \begin{algorithmic}[1]
     
      \REQUIRE $a, b, n_{\text{points}}$
        \STATE $\bm{z} \leftarrow$ evenly spaced $n_{\text{points}}$ numbers over the interval$[a,b]$
        \STATE $c_1, c_2 \leftarrow$ cubic spline coefficients of $\tanh(\bm{z}), \tanh(\bm{z})^2$, where $P_i(\bm{z}) = -c_{i,0}\bm{z}^3 + c_{i,1}\bm{z}^2 - c_{i,2}\bm{z} + c_{i,3} \hspace{1mm}(i = 1,2)$
        \FOR {$j=0$ to $n_{\text{points}}-1$}
        \STATE $\alpha_j, \beta_j \leftarrow \frac{(\bm{z}_j-\mu)}{\sqrt{2\sigma}}, \frac{(\bm{z}_{j+1}-\mu)}{\sqrt{2\sigma}}$
		\STATE $\bm{\mu}_j \leftarrow$ Calculate Eq. \eqref{mean} using Eq. \eqref{spline}
        \STATE $\bm{\sigma}_j \leftarrow$ Calculate Eq. \eqref{variance} using Eqs. \eqref{spline}, \eqref{mean}
        \ENDFOR
	\end{algorithmic}
    \label{alg:spline}
\end{algorithm}
\end{minipage}\hfill
\begin{minipage}[b]{0.48\linewidth}
\begin{table}[H]
\centering
\caption{Typical operation counts for the $tanh(\bm{z})$ approximation}
\label{table: Algorithm complexity analysis}
\subfloat[Spline approximation] {\scalebox{0.7}{
\begin{tabular}
{ccccc}
\hline
\multirow{3}{2cm}{\centering\textbf{Line}} & \multicolumn{4}{c}{\centering\textbf{Algorithm \ref{alg:spline}}}\\
\cline{2-5}
&  $*/\div$ & $+/-$ & $exp$ & $sqrt$\\
\hline
\hline
1 & 2N & 3N &  & \\
2\citep{Press07spline} & 28N & 20N & \\
4 & 4ND & 2ND  &  & 2ND \\
5 & 112ND & 60ND & 18ND & 6ND \\
6 & 112ND+2 & 60ND+1 & 18ND & 6ND \\
\hline
& 228ND+30N+2 & 122ND+23N+1 & 36ND & 14ND \\
\hline
\end{tabular}}}
\quad
\subfloat[Analytical approximation] {\scalebox{0.7}{
\begin{tabular}
{ccccc}
\hline
\multirow{3}{2cm}{\centering\textbf{Eq.}} & \multicolumn{4}{c}{\centering\textbf{Eqs. \eqref{mean1}, \eqref{variance1}}}\\
\cline{2-5}
& $*/\div$ & $+/-$ & $exp$ & $sqrt$ \\
\hline
\hline
Eq. \eqref{mean1} & 6D & 3D & 1D & 1D \\
Eq. \eqref{variance1} & 17D & 7D & 2D & 2D \\
\hline
& 23D & 10D & 3D & 3D \\
\hline
\end{tabular}}}
\end{table}
\end{minipage}
\end{table}

\subsection{Error Bounds for Mean and Variance Estimation}
In this section we provide an upper bound on the mean and variance estimates as a function of the input mean, input variance, the number of nodes in the mesh, and the location of the mesh in space. Let $g(z) \in \mathcal{C}^4$, and define $\tau$ as the maximum interval length in the mesh spanning some compact set $z \in [a, b]$. Then for a cubic spline approximation $P(z)$, we can place an upper bound on the maximum error over the interval $[a,b]$ \citep{de1978practical}. 
\begin{eqnarray}
||g(z) - P(z)||_\infty \leq \frac{1}{16} \tau^4 ||g^{(4)}(z)||_\infty ~,~ z \in [a, b], ~~~~~ \tilde{g}(z) =  \begin{cases}
      -1 &,~~ z \in (-\infty, a) \\
      P(z) &,~~ z\in [a, b] \\
      1 &~,~ z \in (b, -\infty)
     \end{cases}
\end{eqnarray}
Let us use $g(z) = \tanh(z)$. Given the fact that $z \sim \Gaussian{\mu}{\sigma}$, we can compute an error bound over the mean of the output distribution by applying the triangle inequality. 
\begin{eqnarray}
\bigg|\int_{-\infty}^{\infty} (g(z) - \tilde{g}(z))\frac{1}{\sqrt{2 \pi \sigma}} \exp \big({\frac{-1}{2\sigma}(z - \mu)^2}\big) \de z \bigg|
&\leq& c_1 \big( \erf(\frac{b-\mu}{\sqrt{2\sigma}}) - \erf(\frac{a-\mu}{\sqrt{2\sigma}})\nonumber \big) + \\ \nonumber 
&& c_2 \big(  \erf(\frac{a-\mu}{\sqrt{2\sigma}}) + 1 \big) \\
&& + c_3 \big( 1 -  \erf(\frac{b-\mu}{\sqrt{2\sigma}}) \big) = \epsilon_{\mu} \label{int_bound}
\end{eqnarray}
where $c_1 = \frac{1}{32} \tau^4 ||g^{(4)}(z)||_\infty$ , $c_2 = \frac{|1 + \tanh(a)|}{2}$ , $c_3 = \frac{|\tanh(b)-1|}{2}$. Let us additionally compute a bound for the variance. Let $\mu_{\text{true}}$ denote the true value of the mean, $\mu_{\text{spline}}$ be the approximated value of the mean. Additionally, let $\tilde{g}(z)$ be our polynomial approximation of $\tanh(z)^2$. Then we go to the expression for variance Equation \eqref{variance}:
\begin{eqnarray}
\bigg| \big(\E( (\tanh z)^2) -  \mu_{\text{true}}^2 \big) - \big( \E( \tilde{g}(z)) -  \mu_{\text{spline}}^2 \big) \bigg| &\leq& \bigg| \E( (\tanh z)^2 - \tilde{g}(z)) \bigg|  + \bigg|  \mu_{\text{spline}}^2 - \mu_{\text{true}}^2 \bigg| \nonumber \\
&\leq& \epsilon_1  + \bigg|  (\mu_{\text{spline}} - \mu_{\text{true}})(\mu_{\text{spline}} + \mu_{\text{true}}) \label{var_bound1} \bigg| \nonumber \\
&\leq& \epsilon_1 + 2\epsilon_{\mu} = \epsilon_{\sigma}
\end{eqnarray}
$\epsilon_{1}$ is the spline error bound where the unknown function is $\tanh(z)^2$. Additionally, we know that $|\mu_{\text{spline}} - \mu_{\text{true}}| < \epsilon_{\mu}(a, b, \mu, \sigma, \tau)$ and that since we are dealing with the $\tanh$ activation function, the activation mean is upper bounded by 1. Utilizing this method allows us to design the number of uniformly spaced mesh points and the size of the interval according to a desired accuracy. For reference, we utilize the interval [-10, 10] for the mean of the $\tanh$ activation, and use $101$ mesh points, resulting in an error bound of $4.21321\mathrm{e}{-5}$ for the mean of an input distribution with mean $3$ and variance $0.2$. Again, the full derivation for both error bounds can be found in the supplement.
 
\section{Probabilistic Echo State Networks}
The proposed algorithm, Probabilistic Echo State Networks (PESN) derives its internal equations from the deterministic echo state network \citep{jaeger2001echo}. Let $z_k \in \real^{N_x + N_u}$ be input to the network, $y_k \in \real^{N_x}$ be the output of the network, $h_k \in \real^{N_h}$ is the hidden state at time $k$. The deterministic echo state network is characterized by 5 hyperparameters: reservoir size $N_h$, leak rate $L$, noise magnitude $M_e$, sparsity fraction $s$, and spectral radius $r$.
\begin{eqnarray}
h_{k} &=& (1-L) h_{k-1} + L \tanh( W_{\text{in}} z_k + W_{\text{fb}} y_{k-1} + W h_{k-1} ) \label{hidden state step} + M_e \de w\\
y_k &=& W_{\text{out}} \begin{bmatrix} 1 & z_k^T & h_k^T \end{bmatrix}^T \label{readout}
\end{eqnarray}
Here $\de w$ is Gaussian noise with variance 1. The sparsity fraction refers to the fraction on nonzero elements in the matrix $W$, while the spectral radius parameter is the spectral radius of $W$. For the echo state network, we randomly generate the internal weight matrices $W_{\text{in}}$, $W_{\text{fb}}$, and $W$, and we only train the readout weight matrix $W_{\text{out}}$. In our case, training is accomplished using batch least squares. At test time, we recursively update the readout matrix using recursive least squares regression. \citep{simon2006optimal}

For probabilistic echo state networks, assume that the input to the networks are Gaussian. We use moment matching to propagate the Gaussian uncertainty through the nonlinearity in the network, thus we retrieve a Gaussian output.  Our notation is as follows: a Gaussian random vector $\bm{z}$ has a mean $\mu_{\bm{z}}$ and covariance matrix $\Sigma_{\bm{z}}$. We assume that all Gaussian vectors in our echo state network are additionally jointly Gaussian, so that linear combinations of these Gaussian vectors produce another Gaussian vector with appropriate mean vector and covariance matrix. Define the input as $\bm{z}(k)$, the previous target as $\bm{y}(k-1)$ and the hidden state as $\bm{h}(k-1)$.
\begin{eqnarray}
\mu_{\bm{a}} &=& W_{\text{in}} \mu_{\bm{z}(k)} + W_{\text{fb}} \mu_{\bm{y}(k-1)} + W \mu_{\bm{h}(k-1)} \label{input_mu}\\
\Sigma_{\bm{a}} &=& W_{\text{in}} \Sigma_{\bm{z}(k)} W_{\text{in}}^T + W_{\text{fb}} \Sigma_{\bm{y}(k-1)} W_{\text{fb}}^T + W \Sigma_{\bm{h}(k-1)} W^T \label{input_sig}\\
\nonumber && + 2 W_{\text{in}} \Sigma_{\bm{z}(k) \bm{y}(k-1)} W_{\text{fb}}^T + 2W_{\text{fb}} \Sigma_{\bm{y}(k-1) \bm{h}(k-1)} W^T + 2W \Sigma_{\bm{h}(k-1) \bm{z}(k)} W_{\text{in}}^T
\end{eqnarray}
Using our element-wise approximation, we drop the off-diagonal terms in $\bm{a}$ and compute the approximate mean $\mu_{\tanh(\bm{a})}$ and variance $\Sigma_{\tanh(\bm{a})}$ using Equations \eqref{mean} and \eqref{variance}. We also ignore the cross correlation terms, which will cause underestimation of uncertainty.
\begin{eqnarray}
\mu_{\bm{h}(k)} &=& (1-L)\mu_{\bm{h}(k-1)} +  L \mu_{\tanh(\bm{a})} \label{leaky_hidden mu}  \\
\Sigma_{\bm{h}(k)} &=& (1-L)^2\Sigma_{\bm{h}(k-1)} + L^2\Sigma_{\tanh(\bm{a})} + 2(1-L)L\Sigma_{\tanh(\bm{a}) \bm{h}(k-1)} + M_e \Id \label{leaky_hidden sig} \\
\bm{b} &=& \begin{bmatrix} 1,&\bm{z}^T,&\bm{h}^T \end{bmatrix}^T \\
\mu_{\bm{y}(k)} &=&  W_{\text{out}} \mu_{\bm{b}} \label{predict_y mu} \\
\Sigma_{\bm{y}(k)} &=& W_{\text{out}} \Sigma_{\bm{b}} W_{\text{out}}^T \label{predict_y sig}
\end{eqnarray}
Here $\big(N_h, L, M_e, s, r\big)$ are the five hyperparameters for the echo state network, $z$ and $y$ are input and output data, generated or collected from experiments, $n_{\text{train}}$ and $n_{\text{test}}$ are the lengths of the train and test data, $\lambda$ is a forgetting factor, $P$ is $\delta \Id$ where $\Id$ is the identity matrix. Algorithm \ref{PES train} is used for training. Algorithm \ref{PES prediction} is used for single and multi-step prediction for the probabilistic echo state network.

\begin{table}[ht]
\begin{minipage}[b]{0.48\linewidth}
\centering
\begin{algorithm}[H]
  \caption{Training the PESN}
  \label{PES train}
    \begin{algorithmic}[1]
      \scriptsize 
      \REQUIRE $ N_h, L, M_e, s, r, z_{\text{train}}, y_{\text{train}}, N_{\text{washout}}$
        \STATE Randomly initialize $W, W_{\text{in}}$ and $W_{\text{fb}}$
        \STATE Make some elements of $W$ zero to have sparsity fraction of $s$
        \STATE $W \leftarrow W * r / max(eigenvalue(W))$
        \STATE Initialize hidden state with Gaussian distribution
        \FOR{$k=1$ to $n_{\text{washout}}$}
        \STATE Calculate $\bm{\mu}_{h}$ using Eq. \eqref{leaky_hidden mu}
		\ENDFOR
		\FOR{$k=1$ to $n_{\text{train}}$}
        \STATE Calculate $\bm{\mu}_{h(k)}$ using Eq. \eqref{leaky_hidden mu}
		\ENDFOR
        \STATE $Z_{\text{train}} \leftarrow [1, z_{\text{train}}^T, \bm{\mu}_h^T]^T$
        \STATE $W_{\text{out}} \leftarrow y_{\text{train}}Z_{\text{train}}^T\big(Z_{\text{train}}Z_{\text{train}}^T\big)^{-1}$
  \end{algorithmic}
\end{algorithm}
\end{minipage}\hfill
\begin{minipage}[b]{0.48\linewidth}
\centering
\begin{algorithm}[H]
  \caption{PESN single/multi step prediction }
  \label{PES prediction}
    \begin{algorithmic}[1]
      \scriptsize 
      \REQUIRE $ W_{\text{out}}, W_{\text{in}}, N_h, L, M_e, z_{\text{test}}, y_{\text{test}}, \lambda, P$
      	\FOR{$k=1$ to $n_{\text{washout}}$}
        \STATE Calculate $\bm{\mu}_{h}$ using Eq. \eqref{leaky_hidden mu}
		\ENDFOR
		\FOR{$k=1$ to $n_{\text{test}}$}
		\STATE Calculate $\bm{\mu}_{h(k)}$ and $\bm{\sigma}_{h(k)}$ using Eqs. \eqref{input_mu} - \eqref{leaky_hidden sig}
        \STATE $ B \leftarrow [1, z^T_{\text{test}}(k), \bm{\mu}_{h(k)}^T]^T $
		\STATE Calculate $\bm{\mu}_{y(k)}$ and $\bm{\sigma}_{y(k)}$ using Eqs. \eqref{predict_y mu} and \eqref{predict_y sig}
        \STATE Incrementally update $W_{\text{out}}$ and P
        \IF {single step prediction}
        \STATE $y(k) \leftarrow y_{\text{test}}(k)$
        \ELSE 
        \STATE $y(k) \leftarrow \bm{\mu}_{y(k)}$
        \ENDIF
        \ENDFOR
  \end{algorithmic}
\end{algorithm}
\end{minipage}
\end{table}
\section{Experimental Evaluation}
\subsection{Numerical Analysis of Moment Approximation}
First we compare the absolute error between the mean and variance estimated from Monte Carlo against the mean and variance from spline method and analytic method. We evaluate our moment propagation on a range input means and select variance values to assess accuracy. At each point, we have an input Gaussian distribution determined by a mean and variance. We sample this input distribution and compute an estimated mean and variance through Monte Carlo, then compare our two methods. Figure \ref{SvA Moments} shows that the spline method results in a more accurate approximation, and as the input saturates (mean approaches 5 and -5), the error quickly drops. In general we see the analytic approximation over estimates the variance of the $\tanh$ output. Full contour plots of the spline and analytical comparison for a grid of both means and variance are available in the supplementary materials. We see in Figure \ref{fig:err_time} how computational time effects absolute error of moments. The analytic method is constant, but has comparatively high error compared to the others. The spline method has higher initial computational cost, but converges in error quickly.

\begin{figure}[h]
\centering
   \subfloat[Analytical vs Spline: Variance = 0.2] {\includegraphics[scale=0.35]{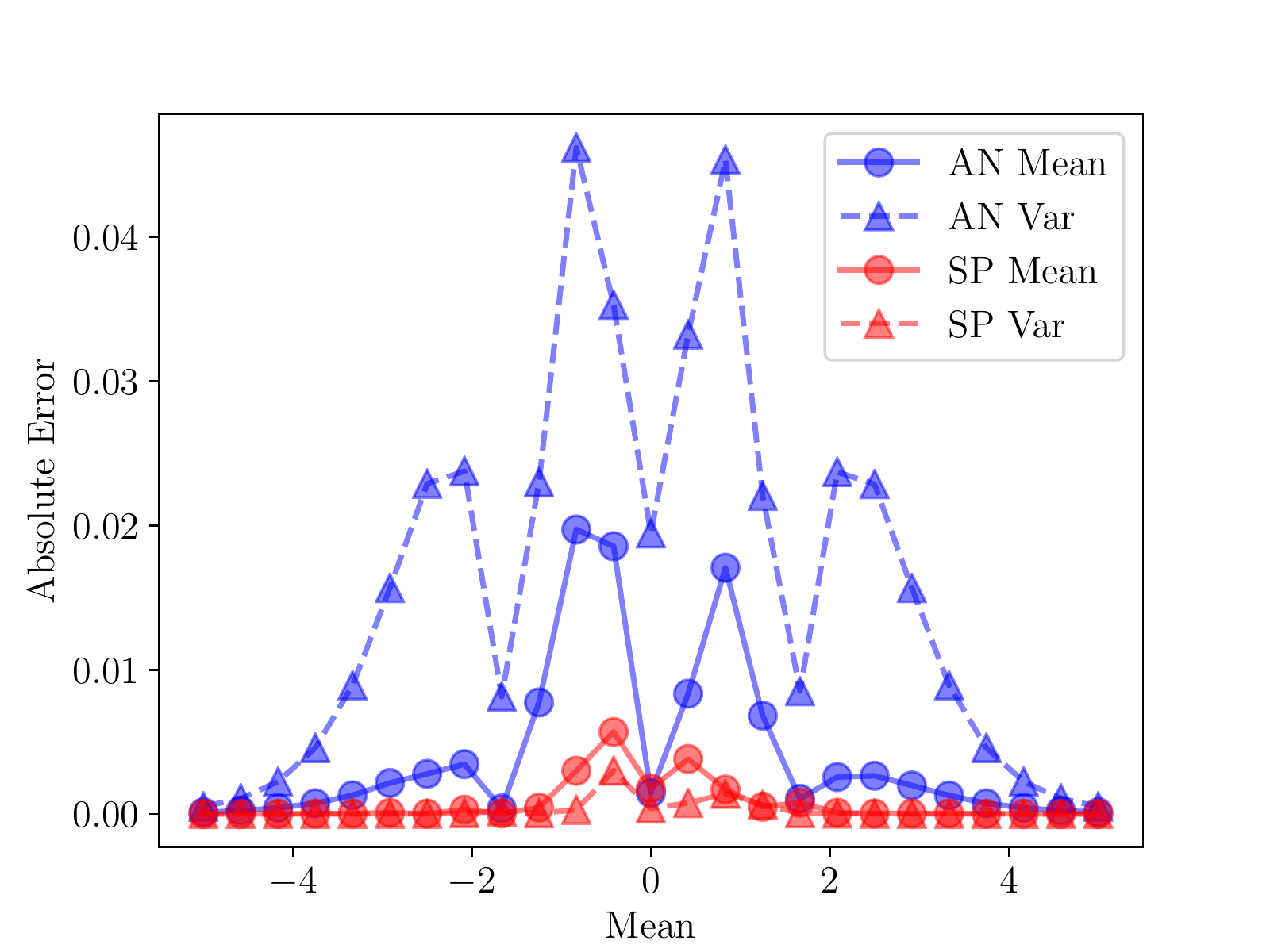}  \label{fig:SPvA Mean}
} ~~~
    \subfloat[Analytical vs Spline: Variance = 1.0] {\includegraphics[scale=0.35]{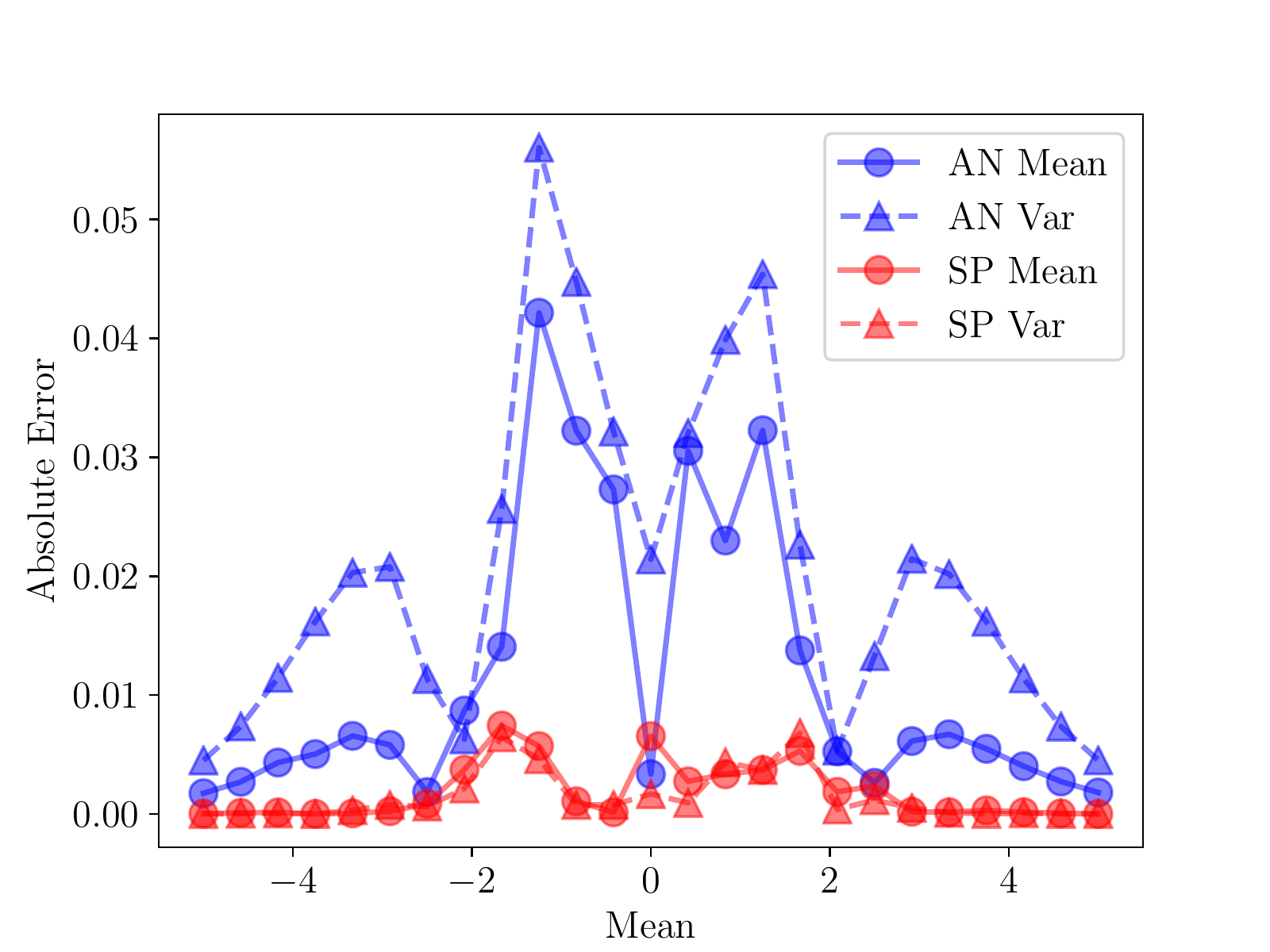}   \label{fig:SPvA Var}
}
\caption{Comparison between the analytic (blue) and spline (red) for small and large variance for mean = -5 to 5. }
\label{SvA Moments}
\end{figure}
\begin{table}[ht]
\begin{minipage}[b]{0.48\linewidth}
\begin{figure}[H]
\centering
\includegraphics[width=0.8\textwidth]{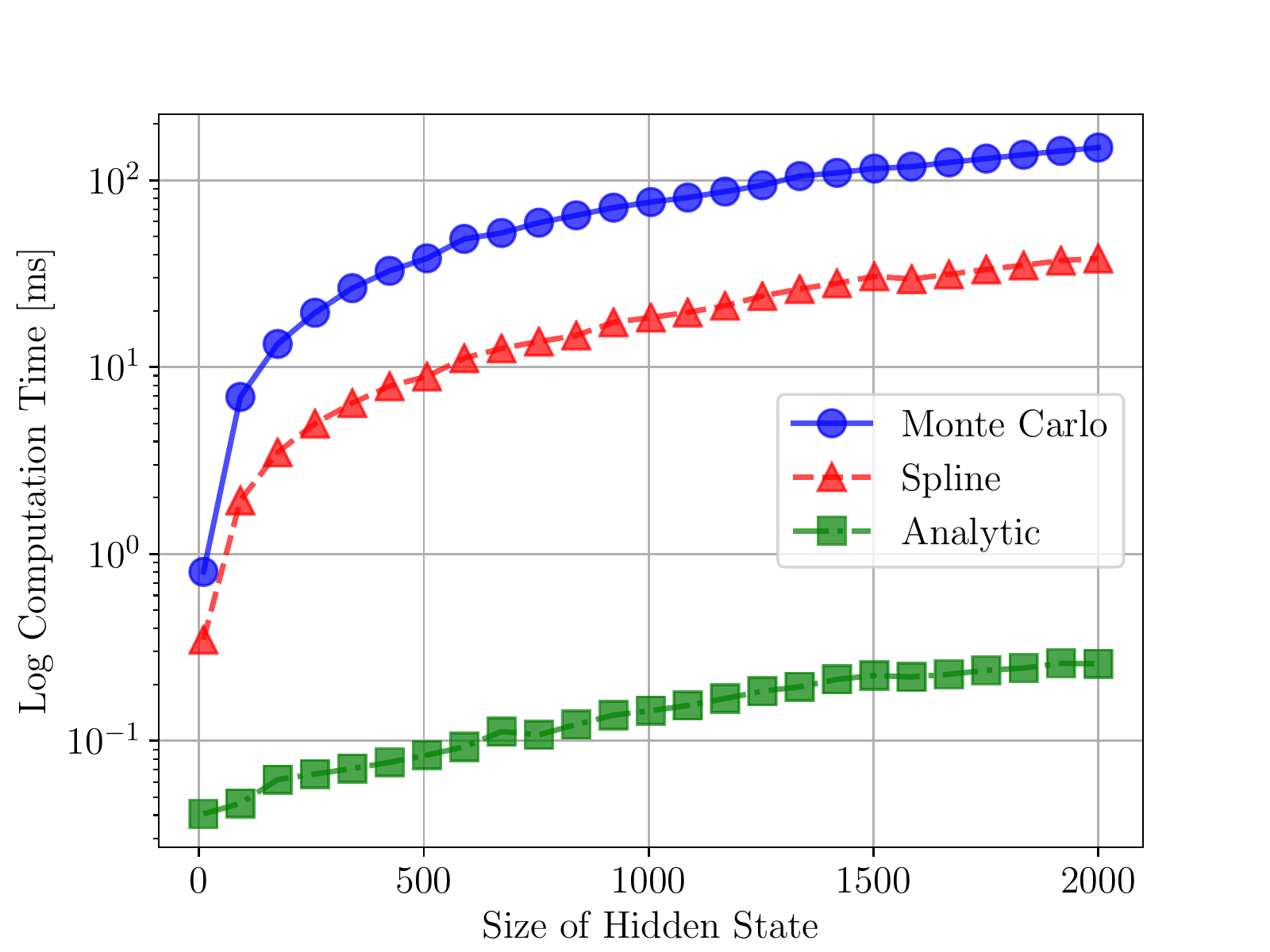}
  \caption{Time Complexity Visualization for Monte Carlo (1e4 Samples), Spline and Analytical Moments}
   \label{fig:time_complexity}
\end{figure}
\end{minipage}\hfill
\begin{minipage}[b]{0.48\linewidth}
\begin{figure}[H]
\centering
 \includegraphics[width=0.8\textwidth]{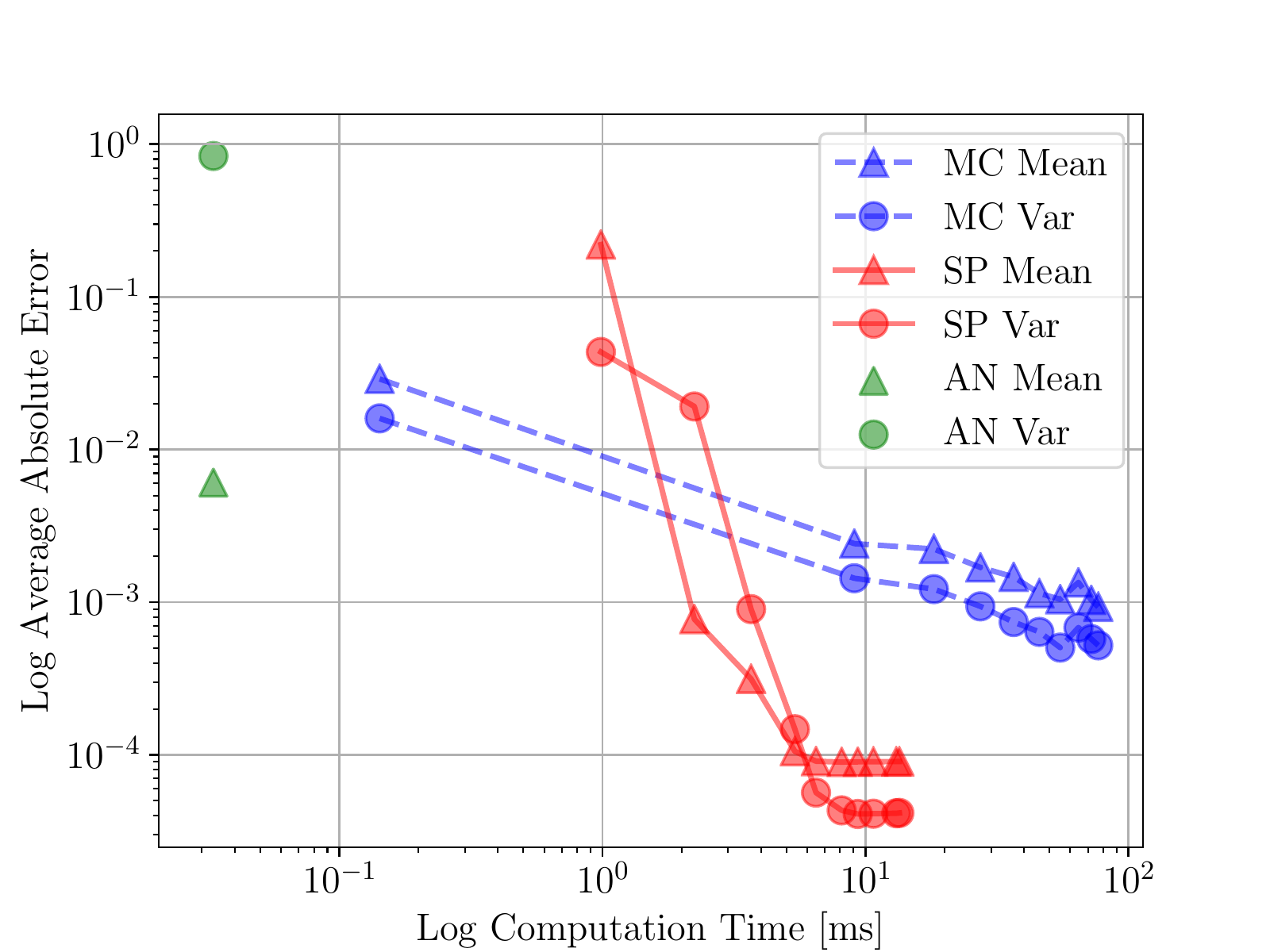}
 \caption{Absolute Error versus Computational Time for Monte Carlo and Spline and Analytical Moments}
   \label{fig:err_time}
   \end{figure}
\end{minipage}
\end{table}
\subsection{Effect of Probabilistic Hidden State on Washout Period}
In Table \ref{table: Washout Comparison} we numerically analyze the difference in multistep prediction error between the probabilistic echo state and the deterministic echo state algorithms as a function of the number of washout points. Each of the 50 trials were run with varying washout lengths (in timesteps), where we ran 50 samples of the deterministic echo state algorithm and compared the average absolute error between the ground truth and predicted state. The table shows the the probabilistic echo state network outperforms the monte-carlo deterministic echo state in absolute error for the majority of washout lengths. When the Monte Carlo deterministic ESN outperforms the PESN, it does so by a fairly small margin. This implies that the PESN reduces the time required to attain the ``echo state property'' that is characteristic for echo state networks. 
\begin{table}
\centering
\caption{Absolute Error Statistics averaged over 50 trials of a 10 timestep trajectory of the Cart Pole. In the ``mean'' column, the minimum value between probabilistic and monte-carlo is bolded. In the ``Washout Length'' column, the minimum error length is bolded.}
\label{table: Washout Comparison}
\subfloat[Position] {\scalebox{0.65}{
\begin{tabular}
{ccccccc}
\hline
\multirow{3}{2cm}{\centering\textbf{Washout Length (timesteps)}} & \multicolumn{6}{c}{\centering\textbf{Error Statistics}}\\
\cline{2-7}
& \multicolumn{2}{c}{\centering\textbf{mean}} 
& \multicolumn{2}{c}{\centering\textbf{min}}
& \multicolumn{2}{c}{\centering\textbf{max}} \\
\cline{2-7}
&  P &MC & P & MC & P & MC \\
\hline
\hline
\textbf{1} &\textbf{ 0.0093} & 0.0128  & 0.0088 & 0.0062  & 0.0134 & 0.0212\\
10 & \textbf{0.0135} & 0.0167  & 0.0134 & 0.0057  & 0.0135 & 0.0377\\
20 & \textbf{0.0428} & \textbf{0.0428}  & 0.0428 & 0.0166  & 0.0429 & 0.0619\\
30 & \textbf{0.0836} & 0.0842  & 0.0836 & 0.0651  & 0.0836 & 0.1007\\
40 & 0.1379 &\textbf{ 0.1359}  & 0.1379 & 0.1139  & 0.1379 & 0.1562\\
50 & \textbf{0.2084} & 0.2091  & 0.2083 & 0.1951  & 0.2084 & 0.2279\\
100 & 0.7138 & \textbf{0.7126}  & 0.7138 & 0.6945  & 0.7138 & 0.7375\\
200 & 0.1525 & \textbf{0.1495}  & 0.1525 & 0.1268  & 0.1525 & 0.1697\end{tabular}}}
\subfloat[Velocity] {\scalebox{0.65}{
\begin{tabular}
{ccccccc}
\hline
\multirow{3}{2cm}{\centering\textbf{Washout Length (timesteps)}} & \multicolumn{6}{c}{\centering\textbf{Error Statistics}}\\
\cline{2-7}
& \multicolumn{2}{c}{\centering\textbf{mean}} 
& \multicolumn{2}{c}{\centering\textbf{min}} 
& \multicolumn{2}{c}{\centering\textbf{max}} \\
\cline{2-7}
&  P &MC & P & MC & P & MC \\
\hline
\hline
1 & \textbf{0.0369} & 0.0401  & 0.0249 & 0.0210  & 0.0381 & 0.0650\\
10 & \textbf{0.0203} & 0.0222  & 0.0201 & 0.0061  & 0.0203 & 0.0462\\
\textbf{20} & \textbf{0.0159} & 0.0191  & 0.0159 & 0.0121  & 0.0159 & 0.0305\\
30 & \textbf{0.0573} & 0.0585  & 0.0573 & 0.0405  & 0.0573 & 0.0793\\
40 & 0.1247 & \textbf{0.1240}  & 0.1247 & 0.0994  & 0.1247 & 0.1402\\
50 & \textbf{0.2206} & 0.2223  & 0.2206 & 0.2078  & 0.2206 & 0.2407\\
100 &\textbf{ 1.3919} & 1.3936  & 1.3919 & 1.3775  & 1.3919 & 1.4133\\
200 & 4.4890 & \textbf{4.4887}  & 4.4890 & 4.4707  & 4.4890 & 4.5095
\end{tabular}}}
\quad
\subfloat[Angle] {\scalebox{0.65}{
\begin{tabular}
{ccccccc}
\hline
\multirow{3}{2cm}{\centering\textbf{Washout Length (timesteps)}} & \multicolumn{6}{c}{\centering\textbf{Error Statistics}}\\
\cline{2-7}
& \multicolumn{2}{c}{\centering\textbf{mean}} 
& \multicolumn{2}{c}{\centering\textbf{min}} 
& \multicolumn{2}{c}{\centering\textbf{max}} \\
\cline{2-7}
&  P &MC & P & MC & P & MC \\
\hline
\hline
\textbf{1} & \textbf{0.3206} & 0.3578  & 0.0653 & 0.3401  & 0.3473 & 0.3808\\
10 & \textbf{0.4380} & 0.4464  & 0.4352 & 0.4230  & 0.4397 & 0.4657\\
20 & \textbf{0.5478} & 0.5496  & 0.5470 & 0.5283  & 0.5485 & 0.5685\\
30 & \textbf{0.6739} & 0.6756  & 0.6734 & 0.6511  & 0.6742 & 0.6955\\
40 & \textbf{0.8240} & \textbf{0.8240}  & 0.8238 & 0.8045  & 0.8243 & 0.8416\\
50 & 1.0001 & \textbf{0.9997}  & 0.9999 & 0.9803  & 1.0002 & 1.0169\\
100 & 0.7372 &\textbf{ 0.7356}  & 0.7371 & 0.7130  & 0.7373 & 0.7562\\
200 & \textbf{0.7244} & 0.7245  & 0.7243 & 0.7053  & 0.7245 & 0.7531
\end{tabular}}}
\subfloat[Angular Velocity] {\scalebox{0.65}{
\begin{tabular}
{ccccccc}
\hline
\multirow{3}{2cm}{\centering\textbf{Washout Length (timesteps)}} & \multicolumn{6}{c}{\centering\textbf{Error Statistics}}\\
\cline{2-7}
& \multicolumn{2}{c}{\centering\textbf{mean}}
& \multicolumn{2}{c}{\centering\textbf{min}} 
& \multicolumn{2}{c}{\centering\textbf{max}} \\
\cline{2-7}
&  P &MC & P & MC & P & MC \\
\hline
\hline
1 & \textbf{0.7104} & 0.7633  & 0.3200 & 0.7406  & 0.7496 & 0.8063\\
10 & \textbf{0.6576} & 0.6652  & 0.6536 & 0.6430  & 0.6601 & 0.6931\\
20 & \textbf{0.5273} & 0.5298  & 0.5260 & 0.5058  & 0.5282 & 0.5597\\
30 & \textbf{0.3817} & 0.3834  & 0.3811 & 0.3599  & 0.3821 & 0.4063\\
\textbf{40 }& \textbf{0.2550} & 0.2558  & 0.2548 & 0.2361  & 0.2552 & 0.2769\\
50 & 0.2761 & \textbf{0.2759}  & 0.2760 & 0.2687  & 0.2761 & 0.2871\\
100 & \textbf{2.8593} & 2.8619  & 2.8592 & 2.8388  & 2.8595 & 2.8920\\
200 & \textbf{2.6587} & 2.6594  & 2.6586 & 2.6350  & 2.6588 & 2.6916\end{tabular}}}
\end{table}
\begin{table}[H]
\centering
\caption{Absolute Error Statistics averaged over 50 trials of a 10 timestep trajectory of the Cart Pole. Lower entropy is bolded}
\label{table: Entropy Comparison}
{\scalebox{0.7}{
\begin{tabular}
{ccccccc}
\hline
\multirow{3}{2cm}{\centering\textbf{Washout Length (timesteps)}} & \multicolumn{6}{c}{\centering\textbf{Shannon Entropy Statistics}}\\
\cline{2-7}
& \multicolumn{2}{c}{\centering\textbf{mean}} 
& \multicolumn{2}{c}{\centering\textbf{min}}
& \multicolumn{2}{c}{\centering\textbf{max}} \\
\cline{2-7}
&  P &MC & P & MC & P & MC \\
\hline
\hline
1 & $\bm{5.42} $& $5.54$ & 5.42 & 5.54  & 5.42 & 5.551\\
10 & $\bm{5.62}  $& $5.69 $ & 5.61 & 5.69  & 5.62 & 5.70 \\
20 & $\bm{5.83} $& $5.86 $ & 5.82 & 5.86  & 5.84 & 5.87 \\
30 & $\bm{5.89}  $& $5.94$ & 5.88 & 5.93  & 5.89 & 5.94 \\
40 & $\bm{5.78}  $& $5.93 $ & 5.77 & 5.92  & 5.78 & 5.93 \\
50 & $\bm{5.62}  $& $5.87 $ & 5.62 & 5.86  & 5.62 & 5.88 \\
100 & $\bm{5.35}  $& $5.71$ & 5.35 & 5.70  & 5.35 & 5.72 \\
200 & $\bm{4.94} $& $5.42$ & 4.94 & 5.41  & 4.95 & 5.45 \end{tabular}}}
\end{table}
In an attempt to quantify the level of synchronization in the hidden states of both the ESN and the PESN, we can create a probability distribution of the values of the reservoir values during washout. Synchronization would imply that there is less uncertainty in the values of the hidden state, thus we can quantify this via the Shannon Entropy of this probability distribution. Table \ref{table: Entropy Comparison} lists the Shannon entropy values for hidden state trajectories. Intuitively, the entropy is small for every washout length, which implies there is less randomness to ``wash out''.
\subsection{Model Learning}
We test the PESN on the task of learning dynamics of cart pole and Gazebo ARDrone. We compare the PESN against Monte Carlo estimates obtained with the deterministic ESN. The PESN successfully propagates uncertainty for single step predictions and for up to 20 timesteps for multi step prediction. The results, found in the supplementary material Section 3.3, demonstrate the ability of our method to capture input uncertainty as it passes through the recurrent network.
\section{Conclusions}
In this work, we investigate two different methods of propagating uncertainty through the $\tanh$ function. We demonstrate that the analytical method is the fastest way to propagate uncertainty through the $\tanh$ function, however it tends to over estimate uncertainty. The spline method is more general and has tuneable accuracy, however the computational cost is higher. Utilizing these new developments, we propose the probabilistic echo state network (PESN), which attempts to solve a fundamental problem of reservoir computing, which is the time required to fulfill the asymptotic behavior of the echo state property. We are able demonstrate, for multi step regression tasks, the PESN has better performance than a deterministic ESN (as measured by absolute error). The better performance is accompanied by a lower shannon entropy in the distribution of hidden state value distribution, which we believe indicates faster convergence to the echo state property. We additionally test the PESN method on ARDrone data collected from the Gazebo simulator, to demonstrate that the method can scale to higher dimensional systems and still propagate uncertainty effectively. However, these methods are not without their drawbacks. For the spline method, input, output scaling and spectral radius of the PESN must be tuned fit inside interval $[a,b]$ so that we satisfy our moment error bounds. The multi-step regression is heavily dependent on incremental updating of the output weights in order to maintain low absolute error. Both the spline and analytic methods are tractable techniques to propagate uncertainty through the $\tanh$ function, and the resulting Probabilistic Echo State network is able to improve the convergence to the echo state property.


\newpage
\bibliography{nips_2018.bib}
\bibliographystyle{unsrtnat} 

\newpage
\section{Supplementary Material}

This is the supplementary material for the paper on Propagating Uncertainty through the Tanh function for Reservoir Computing. The supplement is organized as follows: Section \ref{derivations} includes derivations for both the analytic and spline methods, derivations for the error bound for mean and variance of the spline tanh approximation. Section \ref{PESN} repeats the equations for the Probabilistic Echo State Network (PESN). Section \ref{Experiments} contains additional numerical examples. These include propagation of uncertainty through a feedforward neural network for regression, dynamical systems regression, and activation function comparisons.

\subsection{Propagating Uncertainty through Tanh}
\label{derivations}
\subsubsection{Analytical Approximation of Mean and Variance}
First we present some useful identities and approximations:
The cumulative distribution function (CDF) of a logistic distribution over a random variable $x$ is:
\begin{eqnarray}
F_x(\bm{a}) = \p(x \leq \bm{a}) = \frac{1}{1 + \exp(-\frac{(\bm{a} - \lambda)}{\sigma})} \implies & \p(x \leq 0 | \lambda, \sigma) = \frac{1}{1 + \exp(\frac{\lambda}{\sigma})} \\
\implies & \p(x \geq 0 | \lambda, \sigma) = \frac{1}{1 + \exp(-\frac{\lambda}{\sigma})} \label{logistic cdf}
\end{eqnarray}
Here the logistic distribution is parameterized by the location $\lambda$ and the scale $\sigma$. The $\tanh$ function can be written as follows and relation can be derived:
\begin{eqnarray}
\tanh(\bm{z}) &=& \frac{e^{\bm{z}} - e^{-\bm{z}}}{e^{\bm{z}} + e^{-\bm{z}}} \\
&=& \frac{1 -e^{-2\bm{z}}}{1 + e^{-2\bm{z}}} \\
&=& \frac{1 +1 -(1 + e^{-2\bm{z})} }{1 + e^{-2\bm{z}}} \\
&=& \frac{2}{1 + e^{-2\bm{z}}} - \frac{1 + e^{-2\bm{z}}}{1 + e^{-2\bm{z}}} \\
&=& 2\big(\frac{1}{1 + \exp(-2\bm{z})}\big) - 1 \\
&=& 2 \p(x \geq 0 | \bm{z}, 0.5) - 1 \label{tanh relate}
\end{eqnarray}
Thus the $\tanh$ function is a function of a logistic CDF with location $\lambda = \bm{z}$ (the input to the $\tanh$) and scale $\sigma = 0.5$. Let us assume that the dimensions of the logistically distributed random vector $x$ are independent given the location parameter $\bm{z}$.  From the definition of the logistic distribution, the mean and variance of the conditional distribution $\p(x | \bm{z}, 0.5)$ are:
\begin{eqnarray}
\E(x | \bm{z}) = \bm{z}, ~~~ \Var(z | \bm{z}) = \frac{\pi^2}{12}\Id
\end{eqnarray}
We also make use of the formula for the product of two Gaussian distributions(\cite{petersen2008matrix}).
\begin{eqnarray}
\mathcal{N}_x(\mu_a, \Sigma_a) \times \mathcal{N}_x(\mu_b, \Sigma_b) = c \times \mathcal{N}_x(\mu_c, \Sigma_c) \label{Gaussian prod}
\end{eqnarray}
where
\begin{eqnarray}
c &=& \frac{1}{\sqrt{|2\pi (\Sigma_a + \Sigma_b)|}} \exp\big(-\frac{1}{2}(\mu_a  - \mu_b)^T(\Sigma_a + \Sigma_b)^{-1}(\mu_a  - \mu_b) \big) \\
\mu_c &=& (\Sigma_a^{-1} + \Sigma_b^{-1})^{-1}(\Sigma_a^{-1}\mu_a + \Sigma_b^{-1}\mu_b) \\
\Sigma_c &=& (\Sigma_a^{-1} + \Sigma_b^{-1})^{-1}
\end{eqnarray}
Let us assume that $\bm{z} \sim \Gaussian{\bm{\mu}}{\bm{\sigma}}$. We want to compute the moments of the distribution: $\p(\tanh(\bm{z}))$. We will make use of moment matching to approximate a logistic distribution with a Gaussian, and vice versa.
\begin{eqnarray}
\p(x | \bm{z}, 0.5) \approx x | \bm{z}, 0.5 \sim \Gaussian{\bm{z}}{\frac{\pi^2}{12}\Id} \label{Gaussian Moment Matching}
\end{eqnarray}
\textbf{Mean:} First we compute the mean of $\p(\tanh(\bm{z}))$.
\begin{eqnarray}
\E(\tanh(\bm{z})) &=& \int \tanh(\bm{z}) \p(\bm{z}) \de \bm{z} \\
\text{Eq \eqref{tanh relate}} \implies &=& \int (2 \p(x \geq 0 | \bm{z}, 0.5) - 1)\p(\bm{z}) \de \bm{z} \\
&=& 2\int \p(x \geq 0 | \bm{z}, 0.5)\p(\bm{z}) \de \bm{z} -1 \\
\text{CDF Definition} \implies &=& 2\int \big( \int_0^{\infty} p(x | \bm{z}, 0.5)\de x\big) \p(\bm{z}) \de \bm{z} -1 \\
\text{Eq \eqref{Gaussian Moment Matching}} \implies &\approx& 2\int \big( \int_0^{\infty} \Gaussian{\bm{z}}{\frac{\pi^2}{12}\Id} \de x\big) \p(\bm{z}) \de \bm{z} -1 \\
\text{Distribution of } \bm{z} \implies &=& 2\int \big( \int_0^{\infty} \Gaussian{\bm{z}}{\frac{\pi^2}{12}\Id} \de x\big)  \Gaussian{\bm{\mu}}{\bm{\sigma}} \de \bm{z} -1 \\
\text{Fubini's Theorem} \implies &=& 2 \int_0^{\infty} \big( \int   \Gaussian{\bm{z}}{\frac{\pi^2}{12}\Id}  \Gaussian{\bm{\mu}}{\bm{\sigma}} \de \bm{z} \big)  \de x  -1 \\
\text{Eq \eqref{Gaussian prod}} \implies &=& 2 \int_0^{\infty} \big( \int  (2\pi)^{\frac{n}{2}} |\bm{\sigma} + \frac{\pi^2}{12}\Id|^{u\frac{1}{2}} \p(\bm{z}) \times \label{Prod step} \\
&& \exp(\frac{-1}{2} (x -\bm{\mu})^T ( \bm{\sigma} +  \frac{\pi^2}{12}\Id)^{-1} (x -\bm{\mu}) \de \bm{z} \big)  \de x  -1 \nonumber
\end{eqnarray}
In Equation \eqref{Prod step}, we use the product rule of Gaussians, and specify: $\mu_a = x$, $\mu_b = \bm{\mu}$, $\Sigma_a = \Id \frac{\pi^2}{12}$, and $\Sigma_b = \bm{\sigma}$. Then we are left with the constant coefficient in front of a Gaussian pdf of the input $\bm{z}$.
As a result, the constant coefficient is not a function of the input, thus the integral over this Gaussian pdf integrates to 1.
\begin{eqnarray}
\E(\tanh(\bm{z})) &\approx& 2 \int_0^{\infty} (2\pi)^{\frac{n}{2}} |\bm{\sigma} + \frac{\pi^2}{12}\Id|^{\frac{1}{2}} \exp(\frac{-1}{2} (x -\bm{\mu})^T ( \bm{\sigma} +  \frac{\pi^2}{12}\Id)^{-1} (x -\bm{\mu}) \de x  -1 \nonumber \\
&=& 2 \int_0^\infty \Gaussian{\bm{\mu}}{\bm{\sigma} + \frac{\pi^2}{12}\Id} \de z - 1 \\
&\approx& 2 \p ( x \geq 0 | \bm{\mu},\sqrt{\frac{3}{\pi^2}(~\bm{\sigma} + \frac{\pi^2}{12}\Id)}) - 1 \\
\text{Equation \eqref{logistic cdf}} \implies &=& \frac{2}{1 + \exp(-\frac{\bm{\mu_i}}{\sqrt{\frac{3}{\pi^2}\bm{\sigma_i} + \frac{1}{4}}})} - 1~,~\text{for each element $i$ in $\bm{z}$} \label{mean1}
\end{eqnarray}
Thus, given the input mean and diagonal covariance function, we can compute the posterior mean of an elementwise tanh operation.
\newline
\newline 
\textbf{Variance:} The author in \cite{bassett1998x2} investigated different approximations to the error function. One preliminary solution was the use of the $\tanh$ function to approximate the error function. Using this fact, we see that with the appropriate scaling factors, the Gaussian pdf can be an approximation to the function $(1-\tanh(x))^2$. Again, we are ignoring cross correlation terms, thus the output variance will be diagonal and we can compute each term separately.
\begin{eqnarray}
\text{erf(x)} &\approx& \tanh(\frac{2}{\sqrt{\pi}}x) \\
\implies \frac{\de}{\de x} \text{erf(x)} &\approx& \frac{\de}{\de x} \tanh(\frac{2}{\sqrt{\pi}}x) \\
\implies \frac{2}{\sqrt{2\pi\sigma^2}}\exp(-\frac{y^2}{2\sigma^2}) &\approx& (1-\tanh(y^2)) \label{tanh2 approx}
\end{eqnarray}
where $\sigma^2 = \frac{2}{\pi}$ and $y=\frac{2}{\sqrt{\pi}}x$.
\begin{eqnarray}
\Var(\tanh \bm{z}_i) &=& \E(\tanh( \bm{z}_i)^2) - (\E \tanh(\bm{z}_i))^2 \\
&=& 1 - \E(1 - \tanh(\bm{z}_i)^2) - (\E \tanh(\bm{z}_i))^2 \\
\text{Eq \eqref{tanh2 approx}} \implies &\approx& 1 - \E(\frac{2}{\sqrt{2\pi\sigma^2}}\exp(-\frac{\bm{z}_i^2}{2\sigma^2}) ) - (\E \tanh(\bm{z}_i))^2 \\
&=& 1 - 2 \int \Gaussian{0}{\frac{2}{\pi}} \Gaussian{\bm{\mu}_i}{\bm{\sigma}_{ii}} \de \bm{z}_i - (\E \tanh(\bm{z}_i))^2 \\
\text{Eq \eqref{Gaussian prod}} \implies &=& 1 - 2 \int \frac{1}{\sqrt{2\pi(\frac{2}{\pi} + \bm{\sigma}_{ii})}}\exp(-\frac{\bm{\mu}_i^2}{2(\frac{2}{\pi} + \bm{\sigma}_{ii})} ) \p(\bm{z_i}) \de \bm{z}_i  \\
\nonumber && - (\E \tanh(\bm{z}_i))^2 \label{Gaussian prod step2} \\
&=& 1 - \frac{2}{\sqrt{2\pi(\frac{2}{\pi} + \bm{\sigma}_{ii})}}\exp(-\frac{\bm{\mu}_i^2}{2(\frac{2}{\pi} + \bm{\sigma}_{ii})} ) - \label{variance1} \\ && \nonumber (\frac{2}{1 + \exp(-\frac{\bm{\mu_i}}{\sqrt{\frac{3}{\pi^2}\bm{\sigma_i} + \frac{1}{4}}})} - 1)^2
\end{eqnarray}
%
%


%
\subsubsection{Spline Approximation of Mean and Variance}
In this section, we present a formulation for computing the moments of the $\tanh$ transformation utilizing a spline approximation of the argument of the expectation. Again let us assume that $\bm{z} \sim \Gaussian{\bm{\mu}}{\bm{\sigma}}$, the expectation and variance of the elementwise $\tanh$ are given by the following equations:
\begin{eqnarray}
\E \tanh(\bm{z}_i) &=& \int_{-\infty}^{\infty} \tanh(\bm{z}_i) \frac{1}{\sqrt{2 \pi \sigma_i}} \exp\big({\frac{-1}{2\sigma}(\bm{z}_i - \mu_i)^2}\big) \de \bm{z}_i \label{mean_tanh_spline}\\ \nonumber
\E \big(\tanh(\bm{z}_i) - \E \tanh(\bm{z}_i)\big)^2 &=& 1 - \int_{-\infty}^{\infty} (1 - \tanh(\bm{z}_i)^2) \frac{1}{\sqrt{2 \pi \sigma_i}} \exp\big({\frac{-1}{2\sigma}(\bm{z}_i - \mu_i)^2}\big) \de \bm{z}_i - \\ && -\E \tanh(\bm{z}_i)^2 \label{var_tanh_spline}
\end{eqnarray}
In both cases the integrands are not easily integrable, and depend on the mean and variance of the input. Let us assume that we are dealing with scalar inputs to the activation function, since the vector case is handled elementwise and we are ignoring the off-diagonal elements of the input variance. Instead of direct numeric integration, we propose taking advantage of three facts. First, we know that the tails of the Gaussian pdf approach zero, thus when multiplied against a function with constant tails, the integrand also goes to zero. The $\tanh$, $sigmoid$ both have constant tails. For the $ReLu$ and $swish$ activation functions, the negative tail is 0, and the positive tail is $x$, which has an analytical form when integrated against a Gaussian. Thus we can approximate the integral over the real line using an integral over some compact subset centered around the mean of the input. Second, most activation functions are continuous, and on the compact subset of integration, we can uniformly approximate these functions with polynomials, in fact we will utilize piecewise cubic polynomials for this approximation. Finally, we can derive analytic forms to the integrals of polynomials multiplied against exponentials. The results are expressions for both the output mean and variance that are functions of the exponential function, error function, and input mean and variance. Let $f(z)$ be a scalar valued function that is continuous on the real interval $[a, b]$. Additionally, let $P(z)$ be a piecewise continuous cubic polynomial that interpolates $f$ with $N$ nodes, where each node is the beginning of intervals $[z_j, z_{j+1}]$, where $z_0 = a$ and $z_N = b$.
\begin{eqnarray}
\E f(z) &\approx& \frac{1}{\sqrt{2 \pi \sigma}} \int_{a}^{b} P(z)  \exp\big({\frac{-1}{2\sigma}(z - \mu)^2}\big) \de z \\
&=& \frac{1}{\sqrt{2 \pi \sigma}} \sum_{j=0}^{N-1}  \int_{z_j}^{z_{j+1}} P(z) \exp\big({\frac{-1}{2\sigma}(z - \mu)^2}\big) \de z \\
&=& \frac{1}{\sqrt{2 \pi \sigma}} \sum_{j=0}^{N-1}  \int_{z_j}^{z_{j+1}} \sum_{k=0}^3 c_{jk} z^k  \exp\big({\frac{-1}{2\sigma}(z - \mu)^2}\big) \de z \\
&=& \frac{1}{\sqrt{2 \pi \sigma}} \sum_{j=0}^{N-1} \sum_{k=0}^3 c_{jk}  \int_{z_j}^{z_{j+1}} z^k    \exp\big({\frac{-1}{2\sigma}(z - \mu)^2}\big) \de z \label{spline}
\end{eqnarray}
Within the inner sum, the four terms can be computed analytically, however let us first make a change of variables to simplify the expressions. Let $x = \frac{(z - \mu)}{\sqrt{2\sigma}}$, then $\de z = \sqrt{2\sigma} \de x$. Additionally we should transform our bounds: $\alpha = \frac{(z_j - \mu)}{\sqrt{2\sigma}}$ and $\beta = \frac{(z_{j+1} - \mu)}{\sqrt{2\sigma}}$. Next we can compute the integrals:
\begin{eqnarray}
\frac{1}{\sqrt{2 \pi \sigma}} \int_{z_j}^{z_{j+1}} \exp \big({\frac{-1}{2\sigma}(z - \mu)^2}\big) \de z &=& \frac{1}{\sqrt{\pi}} \int_{\alpha}^{\beta} \exp(-x^2) \de x \\
&=& \frac{1}{2}\big( \erf(\beta) - \erf(\alpha) \big) \label{int_g}
\end{eqnarray}
\begin{eqnarray}
\frac{1}{\sqrt{2 \pi \sigma}} \int_{z_j}^{z_{j+1}} z \exp \big({\frac{-1}{2\sigma}(z - \mu)^2}\big) \de z &=& \frac{1}{\sqrt{\pi}} \int_{\alpha}^{\beta} (x \sqrt{2\sigma} + \mu) \exp(-x^2) \de x \\
&=& \nonumber \sqrt{\frac{\sigma}{2\pi}}\big( \exp(-\alpha^2) - \exp(-\beta^2) \big) + \\
&&  \frac{\mu}{2} \big( \erf(\beta) - \erf(\alpha) \big) \label{int_xg}
\end{eqnarray}
\begin{eqnarray}
\frac{1}{\sqrt{2 \pi \sigma}} \int_{z_j}^{z_{j+1}} z^2 \exp \big({\frac{-1}{2\sigma}(z - \mu)^2}\big) \de z &=& \frac{1}{\sqrt{\pi}} \int_{\alpha}^{\beta} (x \sqrt{2\sigma} + \mu)^2 \exp(-x^2) \de x \\
&=& \nonumber \frac{\sigma}{2} \big( \erf(\beta) - \erf(\alpha) + \frac{2}{/sqrt{\pi}}(\alpha \exp(-\alpha^2) - \beta \exp(-\beta^2)   \big) \\
&& \nonumber + \sqrt{\frac{2\sigma}{\pi}}\big( \exp(-\alpha^2) - \exp(-\beta^2) \big) \\
&&  + \frac{\mu^2}{2} \big( \erf(\beta) - \erf(\alpha) \big) \label{int_x2g}
\end{eqnarray}
\begin{eqnarray}
\frac{1}{\sqrt{2 \pi \sigma}} \int_{z_j}^{z_{j+1}} z^3 \exp \big({\frac{-1}{2\sigma}(z - \mu)^2}\big) \de z &=& \frac{1}{\sqrt{\pi}} \int_{\alpha}^{\beta} (x \sqrt{2\sigma} + \mu)^3 \exp(-x^2) \de x \\
&=& \nonumber \sqrt{\frac{2\sigma^3}{\pi} } \big( (1+\alpha^2)\exp(-\alpha^2) - (1+\beta^2)\exp(-\beta^2)   \big)   \\ 
&& \nonumber \frac{3\sigma \mu}{\sqrt{\pi}} \big( \sqrt{\pi}(\erf(\beta) - \erf(\alpha)) + 2\alpha \exp(-\alpha^2) - 2\beta \exp(-\beta^2)   \big) \\
&& \nonumber + 3\mu^2\sqrt{\frac{\sigma}{2\pi}}\big( \exp(-\alpha^2) - \exp(-\beta^2) \big) \\
&&  + \frac{\mu^3}{2} \big( \erf(\beta) - \erf(\alpha) \big) \label{int_x3g}
\end{eqnarray}
Thus using Equation \eqref{spline}, and Equations \eqref{int_g} - \eqref{int_x3g}, we can analytically compute the expectation of any smooth function $f(z)$ whose tails are constant, or polynomial outside of a compact set. The nodes and coefficients can be computed prior to evaluation of the network, and the integrals are computed once at every layer. The mean and variance the functions to approximate are $f(z)$ and $f(z)^2$ respectively. Both functions have a distinct set of coefficients, but we use the same set of equally spaced nodes for each of computation. Additionally, we can move forward to approximate higher moments of the output distribution and get expressions for the skew and kurtosis, at the cost of having spline approximations for $f(z)^3$ and $f(z)^4$. We can express the mean $\mu$, variance $\sigma$, skewness $\gamma$ and kurtosis $\kappa$.
\begin{eqnarray}
\mu &=& A_1 \label{mean} \\
\sigma &=& A_2 - A_1^2 \label{variance} \\
\gamma &=&  \frac{A_3 - 3 A_1 \sigma - A_1^3}{ \sigma^{3/2}} \label{skew}\\
\kappa &=& \frac{A_4 - 4 A_1 A_3 + 6 A_2^2 A_1^2 - 3 A_1^4}{ \sigma^{2}} \label{kurtosis}
\end{eqnarray}
where
\begin{eqnarray}
A_1 &=& \int_{-\infty}^{\infty} f(\bm{z}_i) \frac{1}{\sqrt{2 \pi \sigma_i}} \exp\big({\frac{-1}{2\sigma}(\bm{z}_i - \mu_i)^2}\big) \de \bm{z}_i \\
A_2 &=& \int_{-\infty}^{\infty} f(\bm{z}_i)^2 \frac{1}{\sqrt{2 \pi \sigma_i}} \exp\big({\frac{-1}{2\sigma}(\bm{z}_i - \mu_i)^2}\big) \de \bm{z}_i \\
A_3 &=& \int_{-\infty}^{\infty} f(\bm{z}_i)^3 \frac{1}{\sqrt{2 \pi \sigma_i}} \exp\big({\frac{-1}{2\sigma}(\bm{z}_i - \mu_i)^2}\big) \de \bm{z}_i \\
A_4 &=& \int_{-\infty}^{\infty} f(\bm{z}_i)^4 \frac{1}{\sqrt{2 \pi \sigma_i}} \exp\big({\frac{-1}{2\sigma}(\bm{z}_i - \mu_i)^2}\big) \de \bm{z}_i
\end{eqnarray}

\textbf{Error Bounds for Spline Mean and Variance Estimation:} In this section we provide an upper bound on the mean and variance estimates as function of the input mean, input variance, the number of nodes in the mesh, and the location of the mesh in space. Let $g(z) \in \mathcal{C}^4$ ( some examples of common activation functions with fourth derivatives are $\tanh$, linear, sigmoid, and swish), and define $\tau$ as the maximum interval length in the mesh spanning some compact set $z \in [a, b]$. Then for a cubic spline approximation $P(x)$, we can place an upper bound on the maximum error over the interval $[a,b]$ \cite{de1978practical}. 
\begin{eqnarray}
||g(z) - P(z)||_\infty \leq \frac{1}{16} \tau^4 ||g^{(4)}(z)||_\infty ~,~ z \in [a, b]
\end{eqnarray}
Let us use $g(z) = \tanh(z)$ as an example to compute an expression for the error for an approximation of $\E \tanh(z)$.
\begin{eqnarray}
\tilde{g}(z) =  \begin{cases}
      -1 &,~~ z \in (-\infty, a) \\
      P(z) &,~~ z\in [a, b] \\
      1 &~,~ z \in (b, -\infty)
     \end{cases}
\end{eqnarray}
Now we approximate expectation of the function $g(z)$, given the fact that $z \sim \Gaussian{\mu}{\sigma}$, by splitting the integral into pieces where the function is approximately constant outside of the interval $[a, b]$, and the integral over $[a, b]$ itself. Then we can bound the approximations of $g(x)$ by their maximum values, pass the constants outside the integral, and evaluate integrals over a Gaussian distribution. Let us define the following:
\begin{eqnarray}
\frac{1}{\sqrt{2 \pi \sigma}} \exp \big({\frac{-1}{2\sigma}(z - \mu)^2} &=& C(z, \mu, \sigma) \\
\implies \int_a^b C(z, \mu, \sigma) &=& \frac{1}{2} \big( \erf(\frac{b-\mu}{\sqrt{2\sigma}}) - \erf(\frac{a-\mu}{\sqrt{2\sigma}}) \big)
\end{eqnarray}

Now we can compute an error bound over the mean of the output distribution by applying the triangle inequality, assumptions about the tail behavior of $\tanh$, our approximation over the interval $[a,b]$. 
\begin{eqnarray}
\bigg|\int_{-\infty}^{\infty} (g(z) - \tilde{g}(z)) C(z, \mu, \sigma) \de z \bigg| &\leq& \bigg|\int_{-\infty}^{a} (g(z) - \tilde{g}(z)) C(z, \mu, \sigma)  \de z \bigg| \nonumber  \\
&& + \bigg|\int_{a}^{b} (g(z) - \tilde{g}(z)) C(z, \mu, \sigma) \de z \bigg| \nonumber \\
&& +\bigg|\int_{b}^{\infty} (g(z) - \tilde{g}(z)) C(z, \mu, \sigma) \de z \bigg| \\ \nonumber
&\leq& \bigg|\int_{-\infty}^{a} |\tanh(a) - (-1)| C(z, \mu, \sigma) \de z \bigg| \\
&& +\bigg|\int_{a}^{b} ||g(z) - \tilde{g}(z)||_\infty C(z, \mu, \sigma)  \de z \bigg| \nonumber \\
&& +\bigg|\int_{b}^{\infty} |\tanh(b) - 1| C(z, \mu, \sigma)  \de z \bigg| \\
&\leq& c_1 \big( \erf(\frac{b-\mu}{\sqrt{2\sigma}}) - \erf(\frac{a-\mu}{\sqrt{2\sigma}})\nonumber \big) \\ \nonumber
&& + c_2 \big(  \erf(\frac{a-\mu}{\sqrt{2\sigma}}) + 1 \big) \\
&& + c_3 \big( 1 -  \erf(\frac{b-\mu}{\sqrt{2\sigma}}) \big) \label{int_bound} \\
&=& \epsilon_{\mu}
\end{eqnarray}
where $c_1 = \frac{1}{32} \tau^4 ||g^{(4)}(z)||_\infty$ , $c_2 = \frac{|1 + \tanh(a)|}{2}$ , $c_3 = \frac{|\tanh(b)-1|}{2}$. We can generalize this bound to any function in the expectation, we simply have to compute the appropriate polynomial coefficients and have an assumption about the tail behavior of the function. For example, let us additionally compute a bound for the variance. Let $\mu_{true}$ denote the true value of the mean, $\mu_{spline}$ be the approximated value of the mean. Additionally, let $\tilde{g}(z)$ be our polynomial approximation of $\tanh(z)^2$. Then we go to the expression for variance Equation \eqref{variance}:
\begin{eqnarray}
\bigg| \big(\E( (\tanh z)^2) -  \mu_{\text{true}}^2 \big) - \big( \E( \tilde{g}(z)) -  \mu_{\text{spline}}^2 \big) \bigg| &\leq& \bigg| \E( (\tanh z)^2 - \tilde{g}(z)) \bigg|  + \bigg|  \mu_{\text{spline}}^2 - \mu_{\text{true}}^2 \bigg| \nonumber \\
&=& \bigg| \E( (\tanh z)^2 - \tilde{g}(z)) \bigg|  + \nonumber \\
&& \bigg|  (\mu_{\text{spline}} - \mu_{\text{true}})(\mu_{\text{spline}} + \mu_{\text{true}}) \label{var_bound1} \bigg|
\end{eqnarray}
In this case, we apply Equation \eqref{int_bound} with the coefficients of the integrand function $g(z) = \tanh(z)^2$, thus we can come up with a bound for the following integral in Equation \eqref{var_bound1}: 
\begin{eqnarray}
\bigg|\int_{-\infty}^{\infty} (\tanh(z)^2 - \tilde{g}(z)) C(z, \mu, \sigma) \big) \de z \bigg| \leq \epsilon_1
\end{eqnarray}
Additionally, we know that $|\mu_{spline} - \mu_{true}| < \epsilon_{\mu}(a, b, \mu, \sigma, \tau)$ and that since we are dealing with the $\tanh$ activation function, the activation mean is upper bounded by 1. Thus:
\begin{eqnarray}
\bigg| \E( (\tanh z)^2 - \tilde{g}(z)) \bigg|  + \bigg|  (\mu_{\text{spline}} - \mu_{\text{true}})(\mu_{\text{spline}} + \mu_{\text{true}}) \label{var_bound1} \bigg| &\leq& \epsilon_1 + 2\epsilon_{\mu} = \epsilon_{\sigma}
\end{eqnarray}
 Utilizing this method allows us to design the number of uniformly spaced mesh points and the size of the interval according to a desired accuracy. For reference, we utilize the interval [-10, 10] for the mean of the $\tanh$ activation, and use $101$ mesh points, resulting in an error bound of $4.21321\mathrm{e}{-5}$ for the mean of an input distribution with mean $3$ and variance $0.2$.
\subsection{Probabilistic Echo State Networks}
\label{PESN}
The proposed algorithm, Probabilistic Echo State Networks (PESN) derives its internal equations from the deterministic echo state network \cite{jaeger2001echo}. Let $z_k \in \real^{N_x + N_u}$ be input to the network, $y_k \in \real^{N_x}$ be the output of the network, $h_k \in \real^{N_h}$ is the hidden state at time $k$. The deterministic echo state network is characterized by 5 hyperparameters: reservoir size $N_h$, leak rate $L$, noise magnitude $M_e$, sparsity fraction $s$, and spectral radius $r$.
\begin{eqnarray}
h_{k} &=& (1-L) h_{k-1} + L \tanh( W_{\text{in}} z_k + W_{\text{fb}} y_{k-1} + W h_{k-1} ) \label{hidden state step} + M_e \de w\\
y_k &=& W_{\text{out}} \begin{bmatrix} 1 & z_k^T & h_k^T \end{bmatrix}^T \label{readout}
\end{eqnarray}
Here $\de w$ is Gaussian noise with variance 1. The sparsity fraction refers to the fraction on nonzero elements in the matrix $W$, while the spectral radius parameter is the spectral radius of $W$. For the echo state network, we randomly generate the internal weight matrices $W_{\text{in}}$, $W_{\text{fb}}$, and $W$, and we only train the readout weight matrix $W_{\text{out}}$. In our case, training is accomplished using batch least squares. At test time, we recursively update the readout matrix using recursive least squares regression. \cite{simon2006optimal}

We now present expressions for probabilistic echo state networks. Here we assume that in that the input to the networks are Gaussian, we use moment matching to propagate the Gaussian uncertainty through the nonlinearity in the network, and we are able to retrieve a Gaussian output.  Our notation is as follows: a Gaussian random vector $\bm{z}$ has a mean $\mu_{\bm{z}}$ and covariance matrix $\Sigma_{\bm{z}}$. We assume that all Gaussian vectors in our echo state network are additionally jointly Gaussian, so that linear combinations of these Gaussian vectors produce another Gaussian vector with appropriate mean vector and covariance vector. That is if $\bm{a}$ and $\bm{b}$ are Gaussian random vectors that are also jointly Gaussian, then: 
\begin{eqnarray}
\bm{c} &=& W_1\bm{a} + W_2\bm{b} \\
\mu_{\bm{c}} &=& W_1\mu_{\bm{a}} + W_2\mu_{\bm{b}} \\
\Sigma_{\bm{c}} &=& W_1\Sigma_{\bm{a}}W_1^T + W_2\Sigma_{\bm{b}}W_2^T + 2W_1\Sigma_{\bm{a} \bm{b}}W_2^T
\end{eqnarray}
Analogously to Equation \eqref{hidden state step} if the input is given by $\bm{z}(k)$, the previous target by $\bm{y}(k-1)$ and the hidden state by $\bm{h}(k-1)$, the argument to the activation function is given by
\begin{eqnarray}
\bm{a} &=& W_{\text{in}} \bm{z}(k) + W_{\text{fb}} \bm{y}(k-1) + W \bm{h}(k-1)\\
\mu_{\bm{a}} &=& W_{\text{in}} \mu_{\bm{z}(k)} + W_{\text{fb}} \mu_{\bm{y}(k-1)} + W \mu_{\bm{h}(k-1)} \label{input_mu}\\
\Sigma_{\bm{a}} &=& W_{\text{in}} \Sigma_{\bm{z}(k)} W_{\text{in}}^T + W_{\text{fb}} \Sigma_{\bm{y}(k-1)} W_{\text{fb}}^T + W \Sigma_{\bm{h}(k-1)} W^T \label{input_sig}\\
\nonumber && + 2 W_{\text{in}} \Sigma_{\bm{z}(k) \bm{y}(k-1)} W_{\text{fb}}^T + 2W_{\text{fb}} \Sigma_{\bm{y}(k-1) \bm{h}(k-1)} W^T + 2W \Sigma_{\bm{h}(k-1) \bm{z}(k)} W_{\text{in}}^T
\end{eqnarray}
Using our element-wise approximation, we drop the off-diagonal terms in $\bm{a}$ and compute the approximate mean $\mu_{\tanh(\bm{a})}$ and variance $\Sigma_{\tanh(\bm{a})}$ using Equations \eqref{mean} and \eqref{variance}.
\begin{eqnarray}
\bm{h}(k) &=& (1-L) \bm{h}(k-1) + L \tanh(\bm{a}) + M_e \de w\label{leaky_hidden} \\
\mu_{\bm{h}(k)} &=& (1-L)\mu_{\bm{h}(k-1)} +  L \mu_{\tanh(\bm{a})} \label{leaky_hidden mu}  \\
\Sigma_{\bm{h}(k)} &=& (1-L)^2\Sigma_{\bm{h}(k-1)} + L^2\Sigma_{\tanh(\bm{a})} + 2(1-L)L\Sigma_{\tanh(\bm{a}) \bm{h}(k-1)} + M_e \Id \label{leaky_hidden sig}
\end{eqnarray}
The leaky hidden state is given by Equation \eqref{leaky_hidden}, then we can construct the readout input $\bm{b} = \begin{bmatrix} 1,&\bm{z}^T,&\bm{h}^T \end{bmatrix}^T$
\begin{eqnarray}
\bm{y}(k) &=& W_{\text{out}} \bm{b} \\
\mu_{\bm{y}(k)} &=&  W_{\text{out}} \mu_{\bm{b}} \label{predict_y mu} \\
\Sigma_{\bm{y}(k)} &=& W_{\text{out}} \Sigma_{\bm{b}} W_{\text{out}}^T \label{predict_y sig}
\end{eqnarray}
Here $\big(N_h, L, M_e, s, r\big)$ are the five hyperparameters for the echo state network, $z$ and $y$ are input and output data, generated or collected from experiments, $n_{\text{train}}$ and $n_{\text{test}}$ are the lengths of the train and test data, $\lambda$ is a forgetting factor, $P$ is $\delta I$ where $I$ is the identity matrix, $W_{\text{in}}, W_{\text{out}}, W_{\text{fb}},$ and $W$ are the weights of the network, $h$ is the hidden state.
\subsection{Experimental Evaluation}
\label{Experiments}
\subsubsection{Activation Function Moment Comparison}
First we compare the absolute error between the mean and variance estimated from Monte Carlo against the mean and variance from spline method and analytic method.
We evaluate our moment propagation on a grid of input means and variances to assess accuracy on a range. At each gridpoint, we have an input Gaussian distribution determined by a mean and variance. We sample this input distribution and compute an estimated mean and variance through Monte Carlo, then compare against our two methods. Figure \ref{SvA Moments} shows that the spline method results in a more accurate approximation, and as the input saturates (mean approaches 5 and -5), the error quickly drops logarithmically. In general we see the analytic approximation over estimates the variance of the $\tanh$ output.
\begin{figure}
\centering
   \subfloat[Absolute Error in Mean] {\includegraphics[scale=0.3]{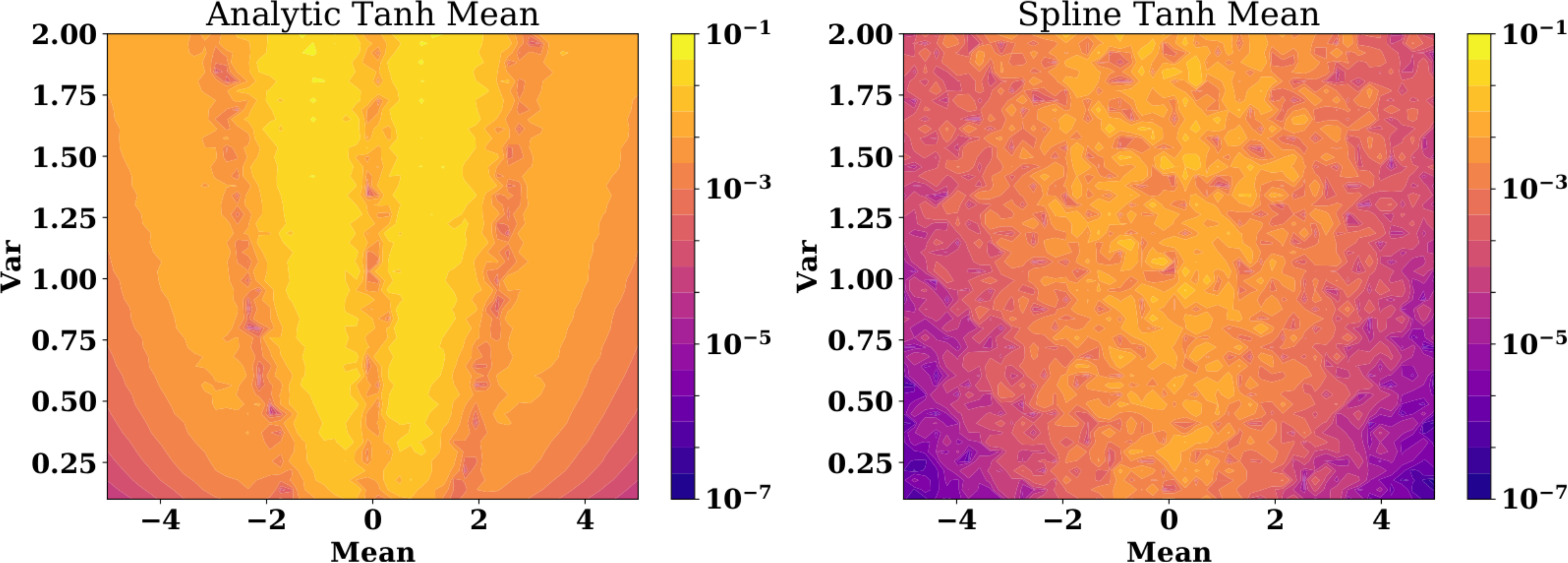}  \label{fig:SPvA Mean}
} 
    \quad
    \subfloat[Absolute Error in Variance] {\includegraphics[scale=0.3]{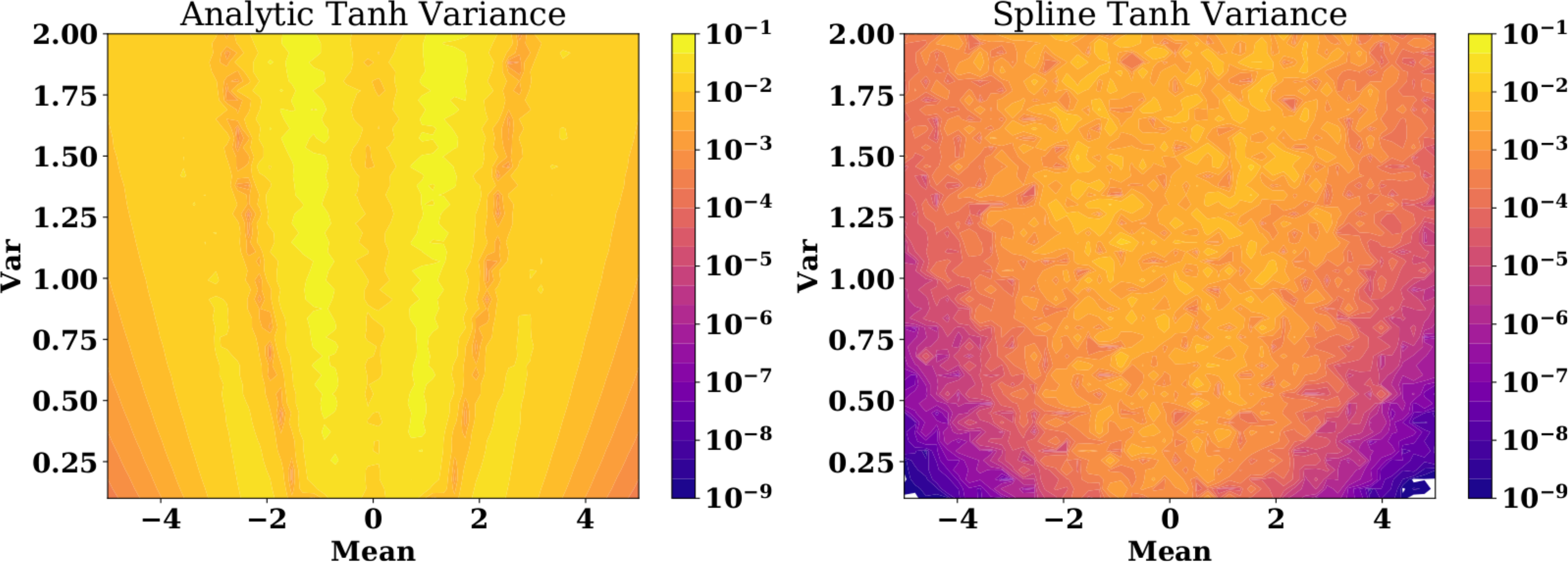}   \label{fig:SPvA Var}
}
\caption{Absolute Error in Moment Comparison between the analytic (left) and spline (right) methods of propagating uncertainty through the $\tanh$. Note that the color bar is in log scale, where darker color implies lower error. }
\label{SvA Moments}
\end{figure}

\begin{figure}
\centering
\subfloat[Tanh] {\includegraphics[scale=0.3]{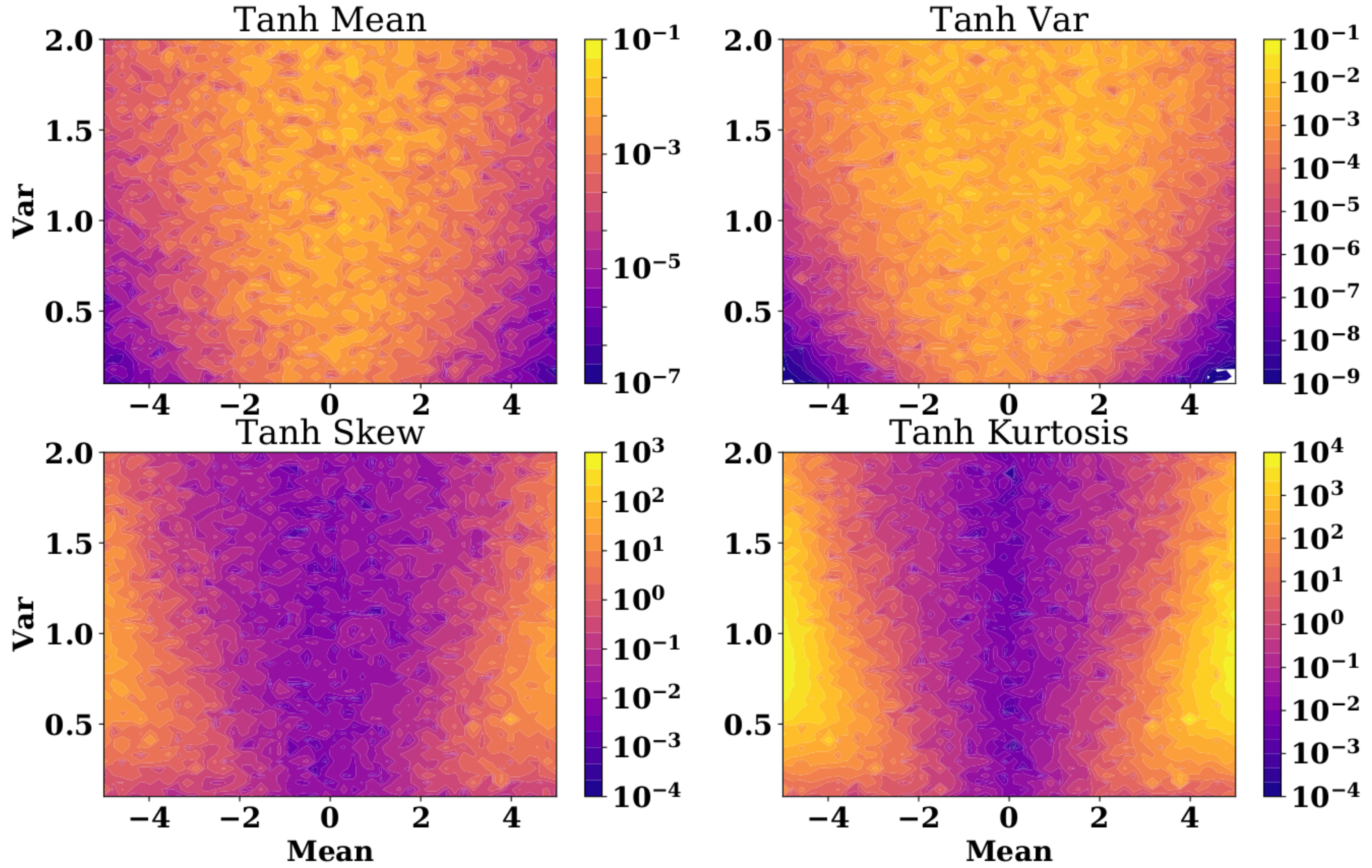} \label{Tanh Moments}} 
\quad
\subfloat[Sigmoid] {\includegraphics[scale=0.3]{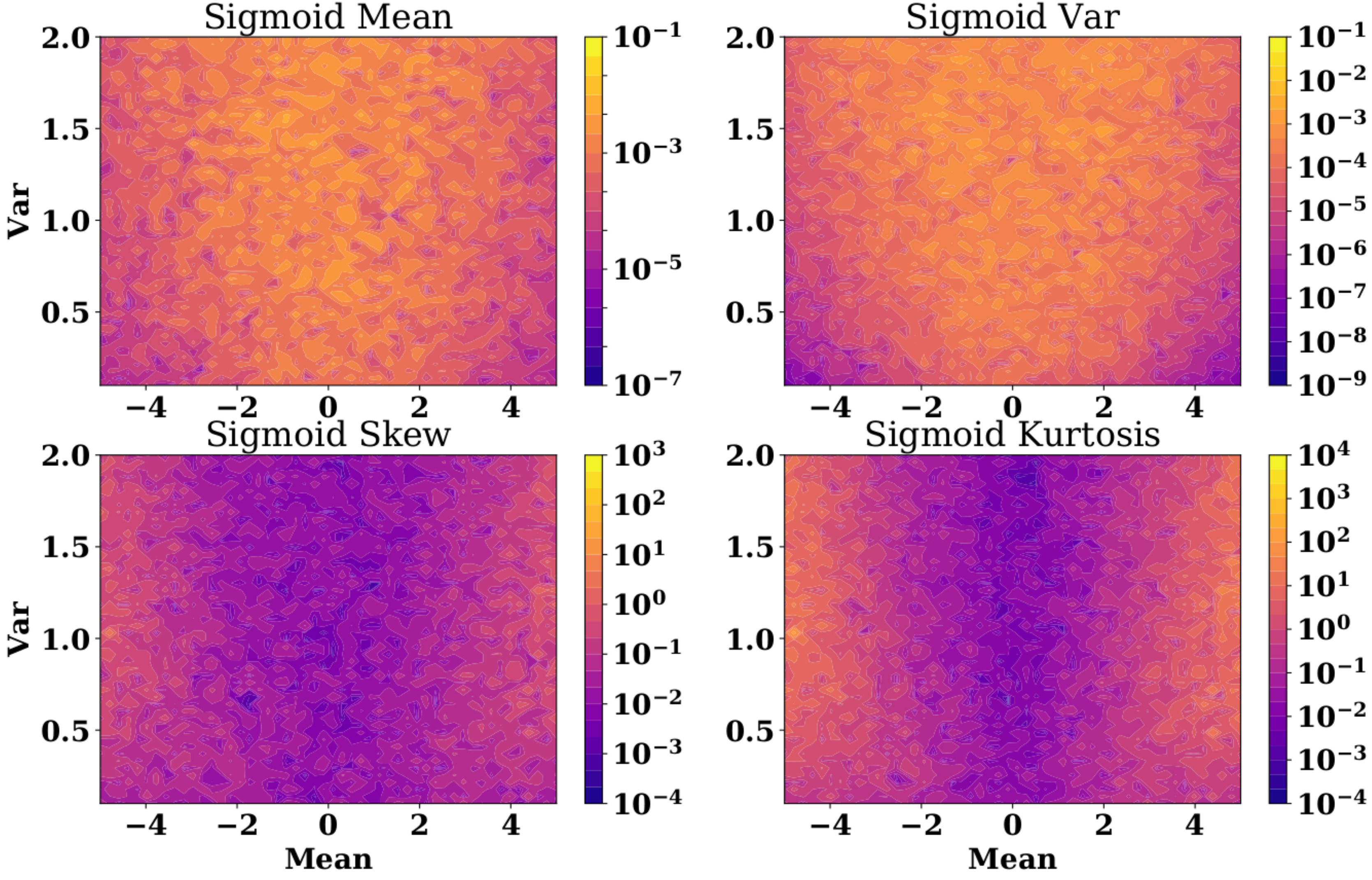} \label{Sig Moments}}
\quad
\subfloat[Swish] {\includegraphics[scale=0.3]{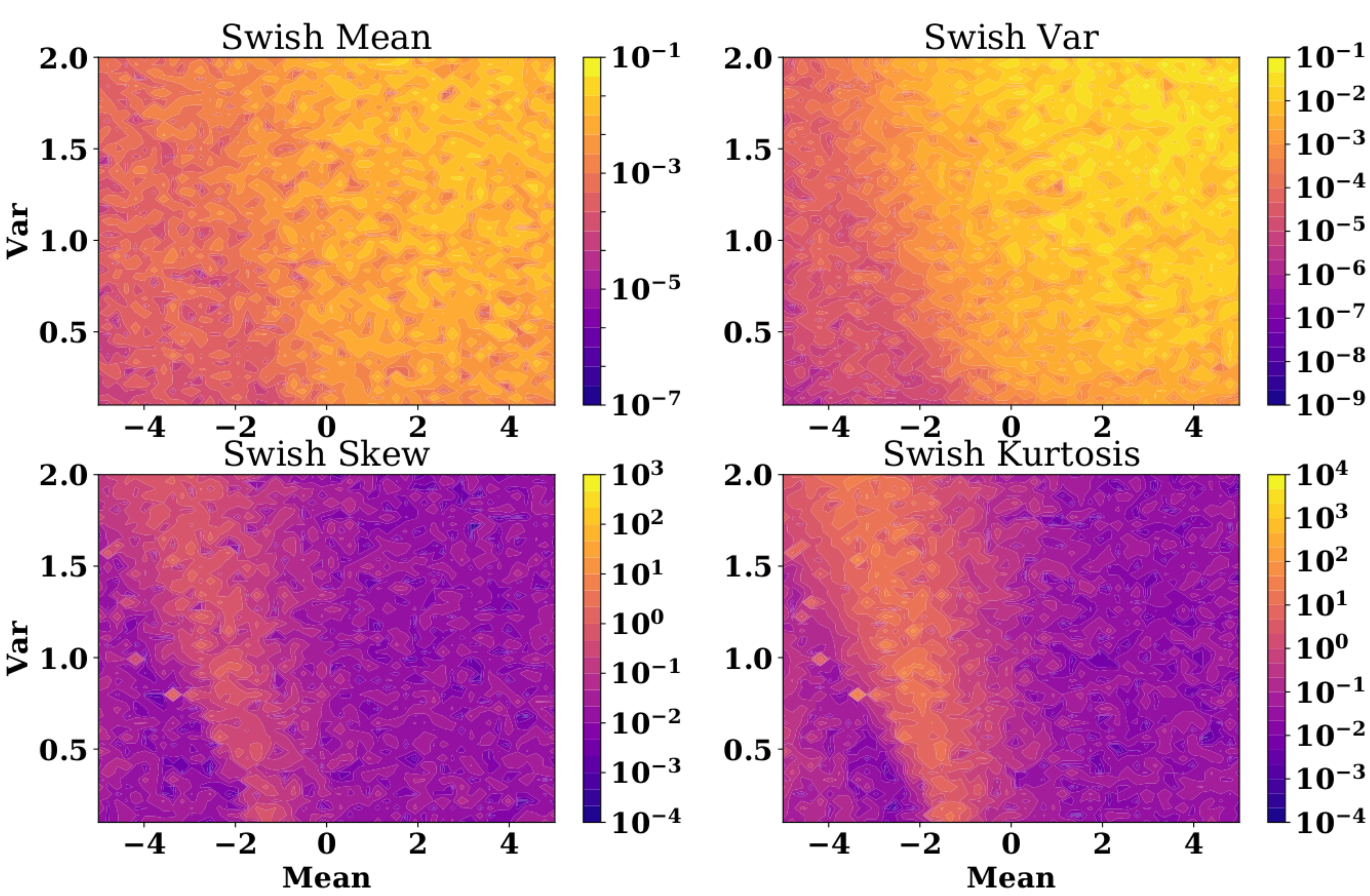} \label{Swish Moments}} 
\caption{Absolute Error in Mean, Variance, Skew and Kurtosis. Again note that the color bar is in log scale, where darker color implies lower error. We have large errors in skew and kurtosis wherever the variance is less than $10e-6$, since these moments are standardized using the variance. }
\label{other moments}
\end{figure}

Additionally, we can compute the skew and kurtosis any continuous activation function. The expressions for these moments are given in Equations \eqref{mean} - \eqref{kurtosis}. We present the absolute error in moments for mean, variance, skew, and kurtosis for 3 activation functions utilizing the spline method: $\tanh$, $sigmoid$, and $swish$ (where $\beta = 1$ for the swish activation function) in Figure \ref{other moments}.
\subsubsection{Multiple Layer Uncertainty Propagation}
In this section, we have constructed two feedforward neural networks with the tanh activation function to demonstrate the performance of the moment approximations. In this case, we randomly initialized all weight matrices. In both cases, the network input is 1024 units and the network output is a single unit. Network 1 has a width of 5 units for 5 hidden layers and network 2 has a width of 50 units and 10 hidden layers. The uncertain input is Gaussian distributed with mean -0.5, and variance 0.01. We see in Table \ref{table:Shallow Network Error} that for network 1, the output is close to a delta function, as a result of the saturation of neurons. The analytical method over estimates the variance in this case, while the spline method is able to recover the moments well. For network 2, shown in Table \ref{table:Deep Network Error}, the distribution goes from a delta function to Gaussian, and both the analytical method and the spline method can recover the output distribution. We see in the table, the absolute error of approximation is smaller for the spline case.
Our error criteria are defined for a given approximation $X$, layer $i$, and number of hidden units $N_{hi}$. $\mu_{MC, i, j}$ is the Monte Carlo approximation mean at layer $i$, hidden unit $j$. The Monte Carlo approximation was computed with 50000 samples.
\begin{eqnarray}
\epsilon_{\mu_X, i} = \frac{1}{N_{hi}}\sum_{j=1}^{N_{hi}}|\mu_{MC, i, j} - \mu_{X, i, j}| \\
\epsilon_{\sigma_X, i} = \frac{1}{N_{hi}}\sum_{j=1}^{N_{hi}}|\sigma_{MC, i, j} - \sigma_{X, i, j}|
\end{eqnarray}
\begin{table}[h]
\centering
\caption{Absolute Error Statistics for FFNN. Subscript ``S'' represents the spline method, while ``A'' represents the analytical method. The input is 1024 units, and the output is 1 unit. The uncertain input is Gaussian distributed with mean -0.5, and variance 0.01.
}
\subfloat[Shallow Network with 5 Hidden Units and 5 Layers] {\scalebox{1.0}{
\begin{tabular}
{c|c|c|c|c|c|c}
\hline
\multirow{2}{2cm}{\centering\textbf{Error Statistics}} & \multicolumn{6}{c}{\centering\textbf{Layers}}\\
\cline{2-7}
&  1 & 2 &  3 & 4 & 5 & Out \\
\hline
\hline
$\mu_S$ & \textbf{2.24e-05} & \textbf{4.09e-05} & \textbf{3.08e-05} & \textbf{5.02e-05} & \textbf{6.02e-05} &\textbf{ 2.04e-05} \\
$\mu_A$ & 8.91e-04 & 8.82e-04 & 9.77e-03 & 1.12e-02 & 1.43e-02 & 6.26e-02 \\
\hline
$\sigma_S$  &\textbf{ 3.31e-05} & \textbf{5.98e-05} & \textbf{3.48e-05} &\textbf{ 2.66e-05} & \textbf{2.04e-05} & \textbf{7.43e-05} \\
$\sigma_A$  & 1.79e-03 & 2.10e-02 & 1.20e-02 & 1.99e-02 & 2.54e-02 & 1.05e-01 
\label{table:Shallow Network Error}
\end{tabular}}}
\quad
\subfloat[Deep Network with 50 Hidden Units and 10 Layers] {\scalebox{1.0}{
\begin{tabular}
{c|c|c|c|c|c|c}
\hline
\multirow{2}{2cm}{\centering\textbf{Error Statistics}} & \multicolumn{6}{c}{\centering\textbf{Layers}}\\
\cline{2-7}
&  1 & 3 &  5 & 7 & 9 & Out \\
\hline
\hline
$\mu_S$ & \textbf{7.97e-04} & \textbf{1.01e-02} & 5.19e-02 & 7.99e-02 & 8.27e-02 & \textbf{9.81e-02} \\
$\mu_A$ & 5.98e-03 & 1.58e-02 & \textbf{5.06e-02} & \textbf{7.48e-02} & \textbf{8.16e-02} & 1.05e-01 \\
\hline
$\sigma_S$  & \textbf{1.03e-03} & \textbf{6.60e-02} & 7.50e-02 & \textbf{7.11e-02} & \textbf{3.90e-02} & \textbf{5.30e-01} \\
$\sigma_A$  & 8.44e-03 & 6.64e-02 & \textbf{7.13e-02} & 9.57e-02 & 7.13e-02 & 2.95e+00
\label{table:Deep Network Error}
\end{tabular}}}
\end{table}
\begin{figure}[h]
\centering
   \subfloat[Shallow Network CDF] {\includegraphics[scale=0.23]{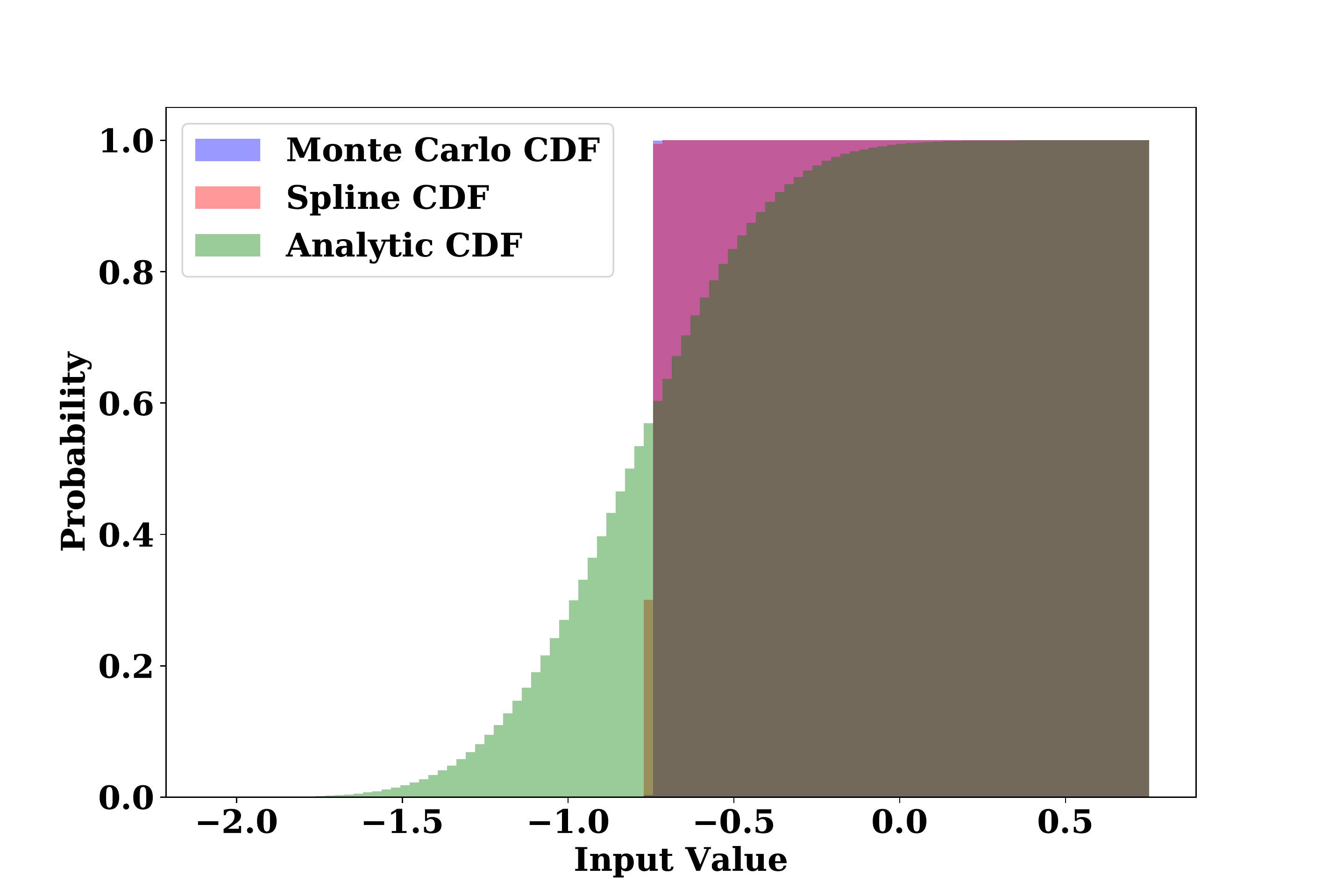}  \label{fig:Shallow CDF}
}  
    \subfloat[Deep and Wide Network CDF] {\includegraphics[scale=0.23]{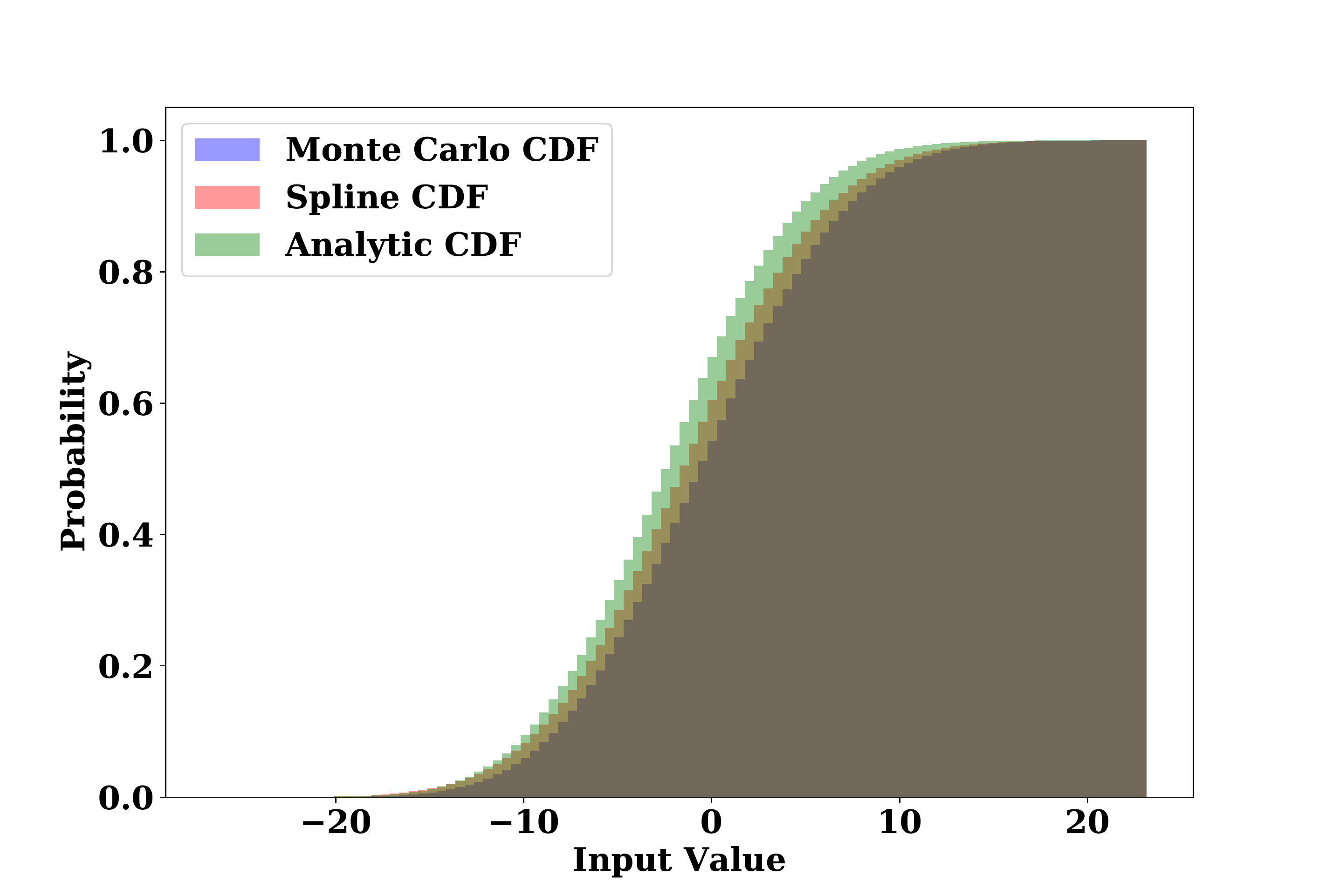}   \label{fig:Deep CDF}
}
\caption{ CDF comparison for two different networks. In the shallow network, the output cdf is close to a delta function. In the deep and wide network, the output cdf is close to a Gaussian distribution.}
\end{figure}
\subsubsection{Model learning}
In the field of robotics and autonomy, recently there is an increasing interest in using machine learning to get the accurate model approximation. Model accuracy is important for many applications including planning and control. However, analytical models cannot catch the complexity of a system very well, so we approximate the system model by learning from data. In this section, we learn unknown system dynamics from data using probabilistic echo state networks(ESN) and evaluate the learned model. We suppose our systems' dynamics are unknown and represent the dynamics as the following differential equation
\begin{equation}
d\bm{x} = f(\bm{x},\bm{u})dt + C(\bm{x},\bm{u})d\omega, \hspace{5mm} d\omega \sim \Gaussian{0}{\Sigma_\omega}
\end{equation}
,where $\bm{x} \in \mathbb{R}^{N_x}$ is the state and $\bm{u} \in \mathbb{R}^{N_u}$ is the control. Given the state-control pair and the state transition $d\bm{x}$, we are able to infer the dynamics as well as the state transition. With the goal of inferring $d\bm{x}$ given $(\bm{x},\bm{u})$, we compare our probabilistic ESN using spline approximation of $tanh$ to Monte Carlo(MC) sampling of the deterministic ESN using $tanh$ under uncertain inputs. To make the inputs uncertain for the deterministic ESN, we added Gaussian noise to the state $\bm{x}$ when we do the MC sampling. For fair comparison, the two networks shared same weights $W_{in}, W_{out}, W_{fb},$ and $W$.
Data for the Cart Pole system was generated by equations of motion derived using first principles and the ARDrone data was collected using a Gazebo simulator \cite{engel12ardrone}. As the kinematics of each system is straightforward and easy to derive, we directly use the kinematics equations to predict next position states and we only learn the dynamics of the systems with ESNs.
\newline
\newline 
\textbf{Cart Pole:} We consider the Cart Pole having a state vector $\bm{x}=[x, \theta, \dot{x}, \omega]^T$, where $x$ is cart position, $\theta$ is pole angle, $\dot{x}$ is cart velocity and $\omega$ is pole angular velocity. The control input to the system $\bm{u}$ is the force applied to the cart. Through several times of simulation with different controls, we collected various data.
We use $[\bm{x}^T, \bm{u}]^T$ as an input to the network. To learn the dynamics of the system, we set the target output as the two velocity terms $[\dot{x}, \omega]^T$.
\begin{figure}
\centering
  \subfloat[Single step prediction] {{\includegraphics[scale=0.11]{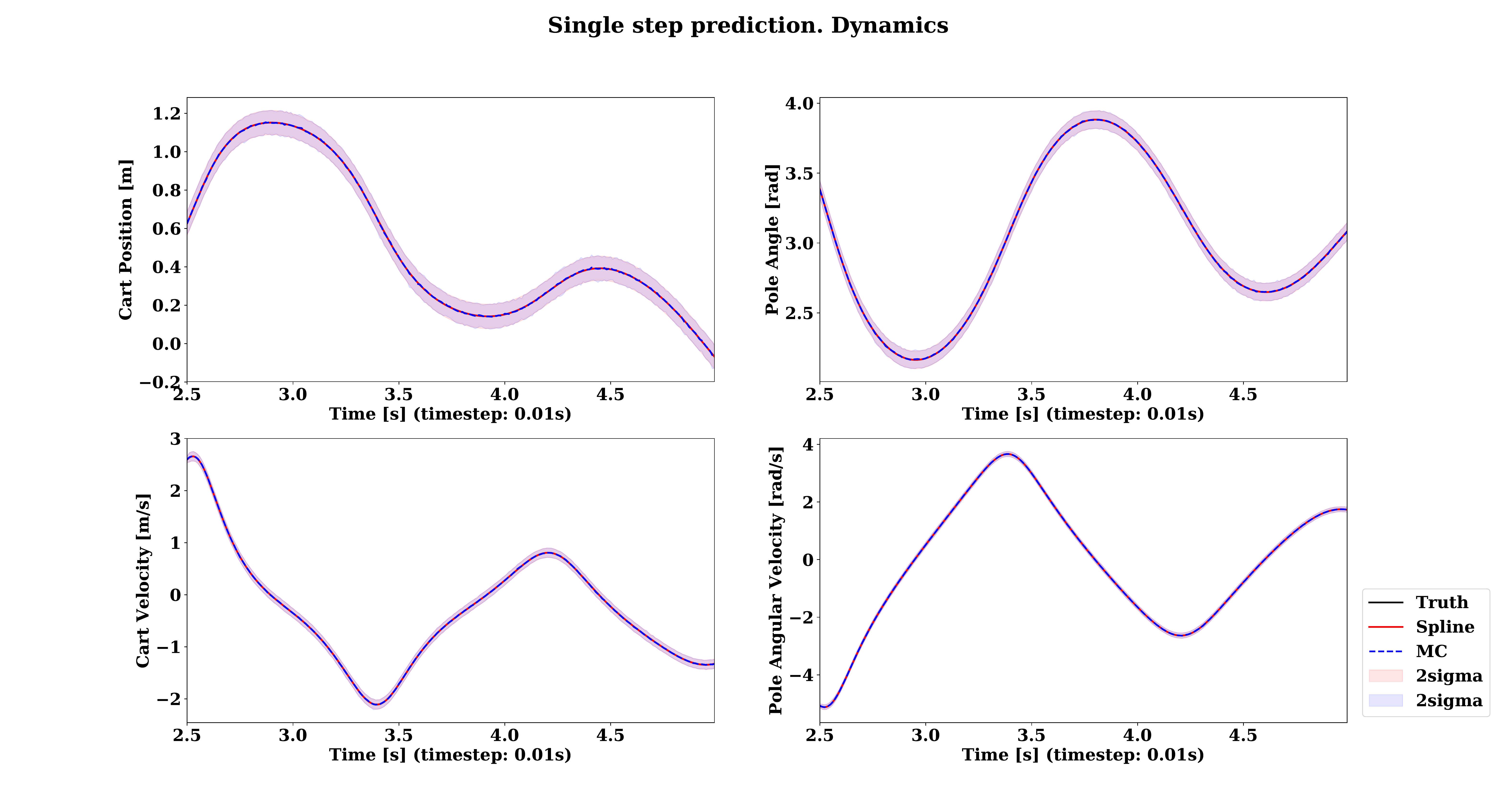} }}
  \subfloat[Zoomed single step prediction] {{\includegraphics[scale=0.11]{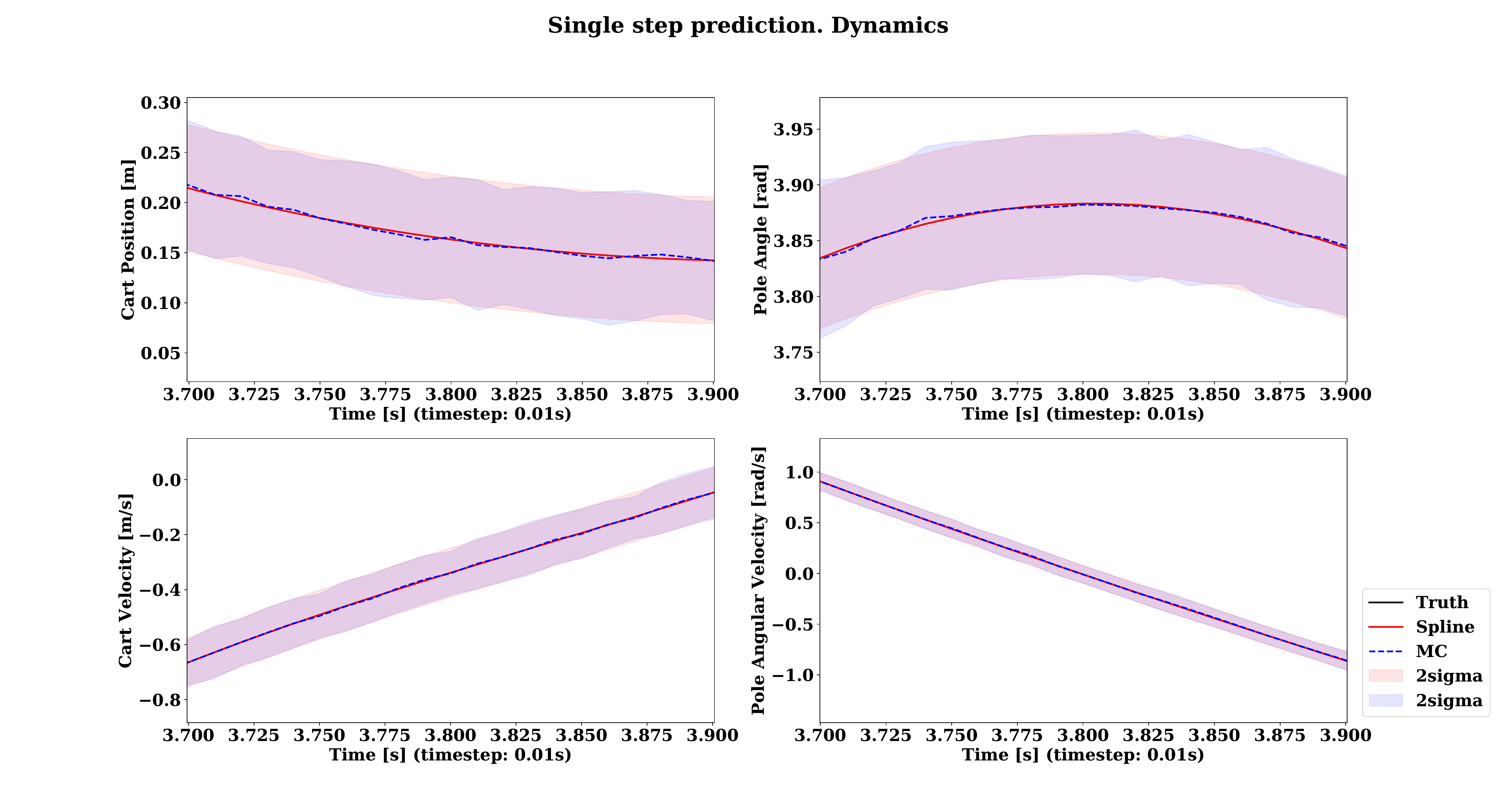} }}
  \quad
  \subfloat[Multi step prediction] {{\includegraphics[scale=0.11]{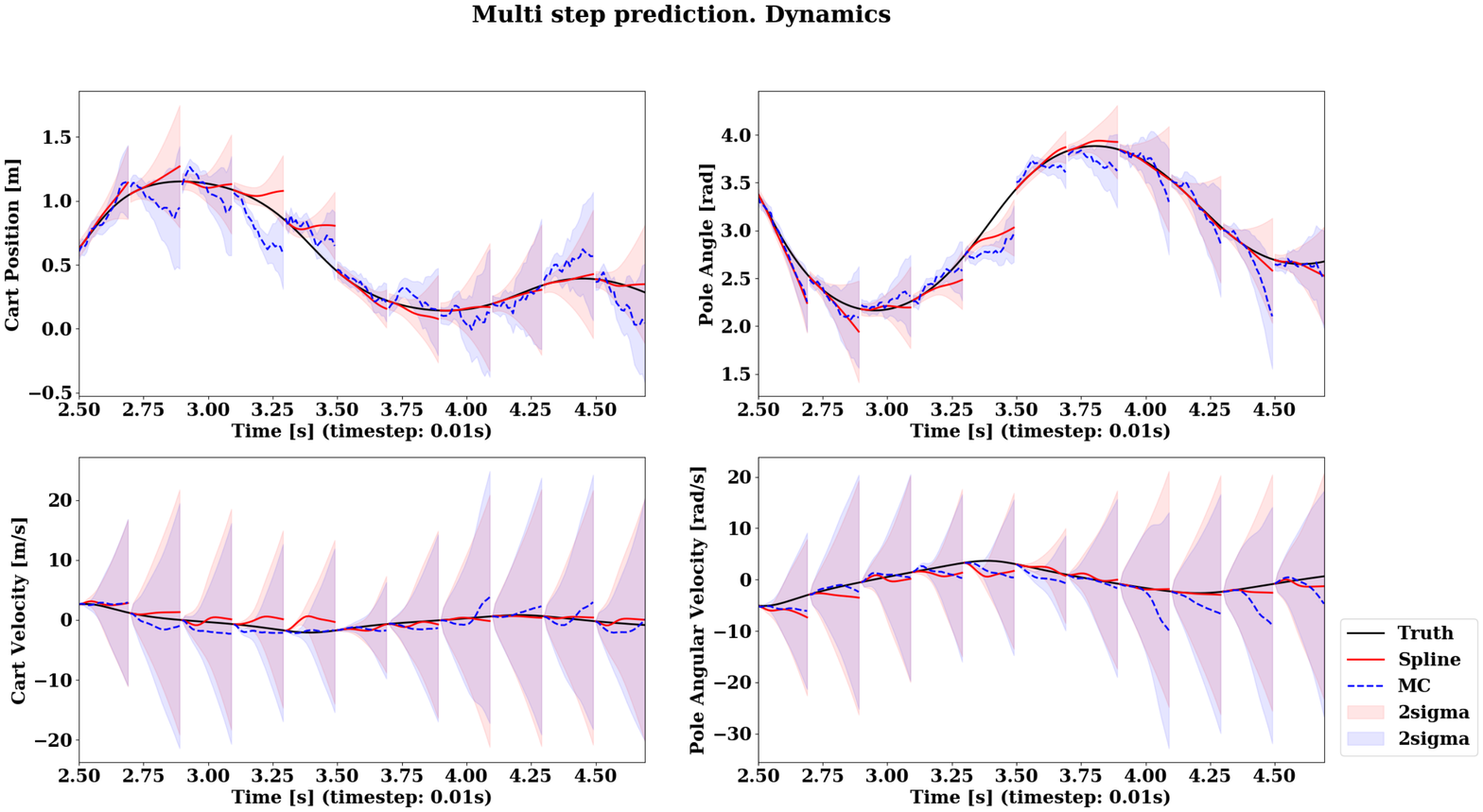} }}
  \subfloat[Zoomed multi step prediction] {{\includegraphics[scale=0.11]{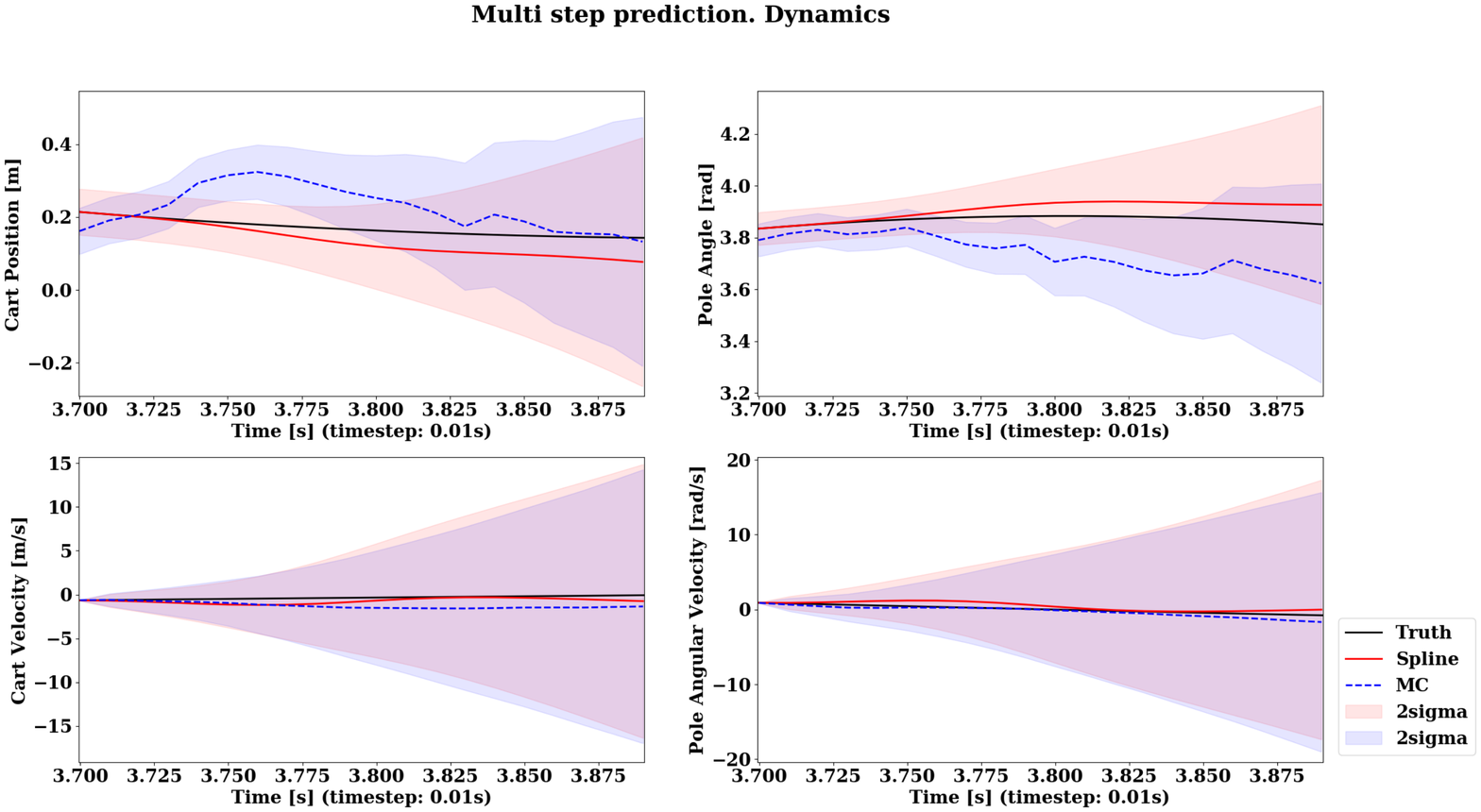} }}
  \quad
  \caption{Cart pole state transition prediction. The red line and shadow represent the predicted mean and 2$\sigma$ (standard deviation) from probabilistic ESN using spline-approximated $tanh$. The blue line and shadow are from MC sampling of deterministic ESN under uncertain inputs. Multi step prediction plots 20 time steps starting from the ground truth(black).}
  \label{fig:CartPole}
\end{figure}

\textbf{ARDrone:}
In the simulated ARDrone, we use 12 state of the system. $\bm{x}=[x, y, z, \dot{x}, \dot{y}, \dot{z}, \phi, \theta, \psi, \dot{\phi}, \dot{\theta}, \dot{\psi}]^T$, where $(x, y, z)$ are positions, $(\phi, \theta, \psi)$ are angles, $(\dot{x}, \dot{y}, \dot{z})$ are linear velocities and $(\dot{\phi}, \dot{\theta}, \dot{\psi})$ are roll rate, pitch rate and yaw rate each. The control input to the ARDrone system is $\bm{u}=[v_{x,cmd}, v_{y,cmd}, v_{z,cmd}, v_{\psi,cmd}]^T$. Here $(v_{x,cmd}, v_{y,cmd}, v_{z,cmd})$ are linear $x,y,z$ velocity commands and $v_{\psi,cmd}$ is the angular $z$ velocity command. Running the simulated ARDrone with various control commands, we collected 80,000+ data. Using
$[\bm{x}^T, \bm{u}^T]^T$ as an input to the network, we set the output target as velocity terms $[\dot{x}, \dot{x}, \dot{y}, \dot{\phi}, \dot{\theta}, \dot{\psi}]^T$ to learn the dynamics of the drone.

\begin{figure}
\centering
  \subfloat[Single step prediction] {{\includegraphics[scale=0.18]{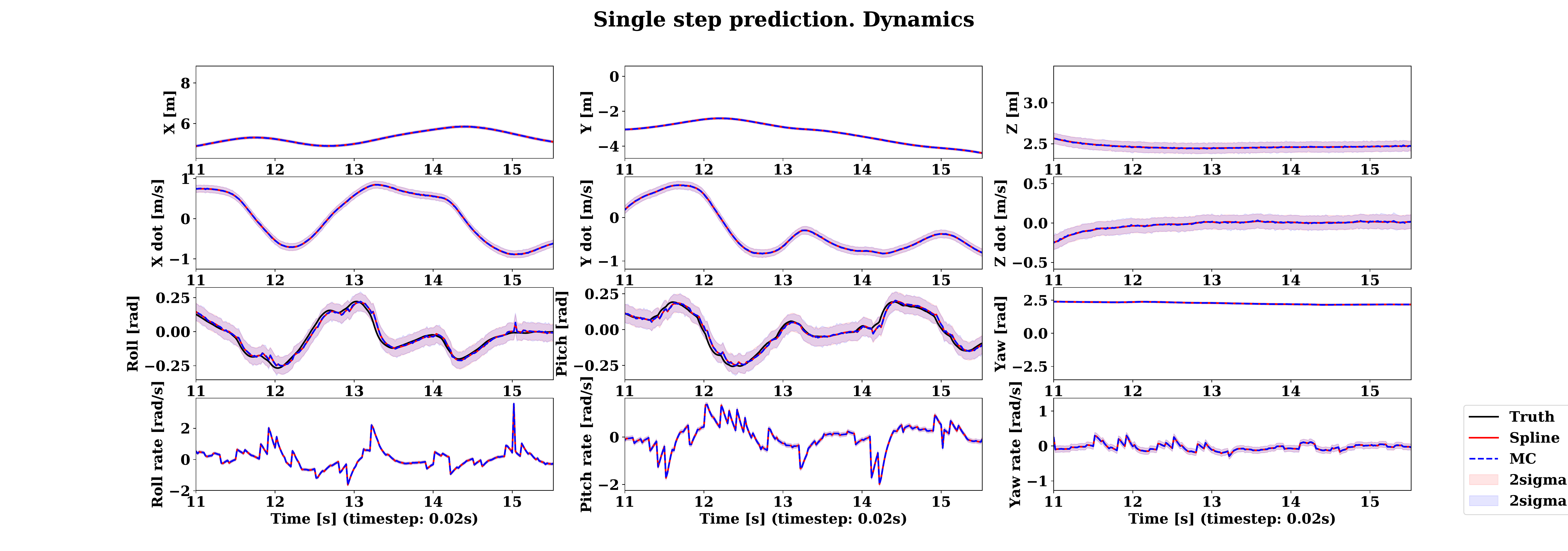} }}
  \quad
  \subfloat[Zoomed single step prediction] {{\includegraphics[scale=0.18]{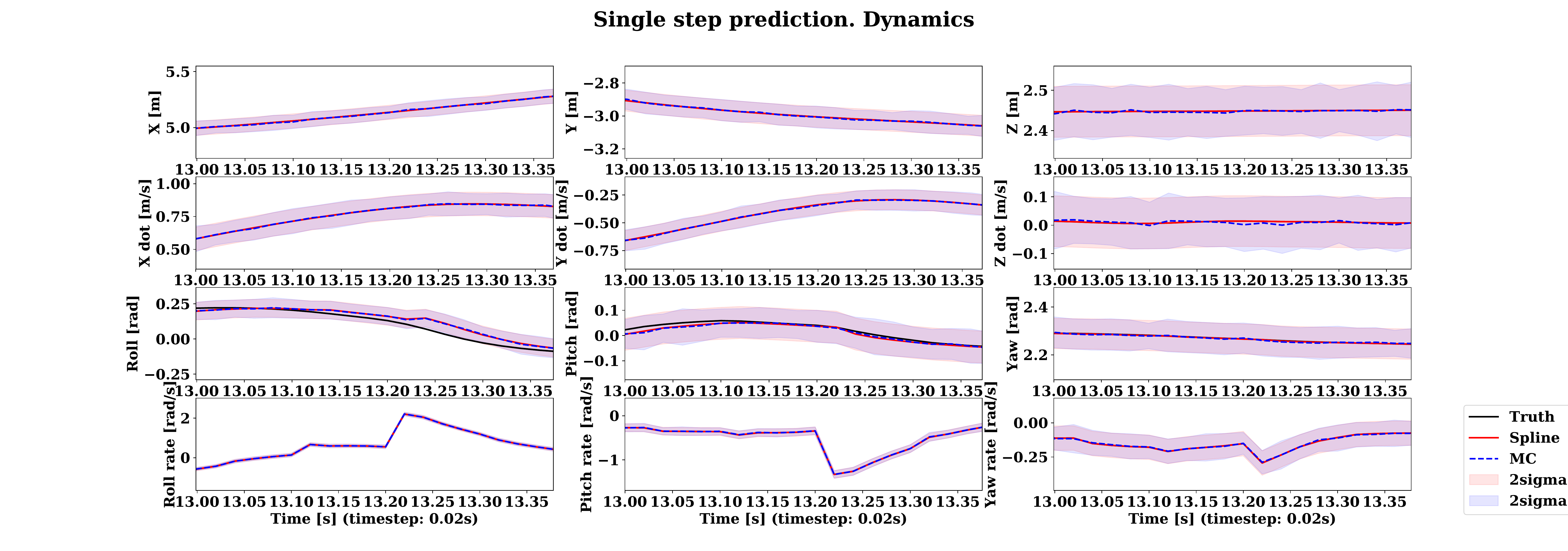} }}
  \quad
  \subfloat[Multi step prediction] {{\includegraphics[scale=0.18]{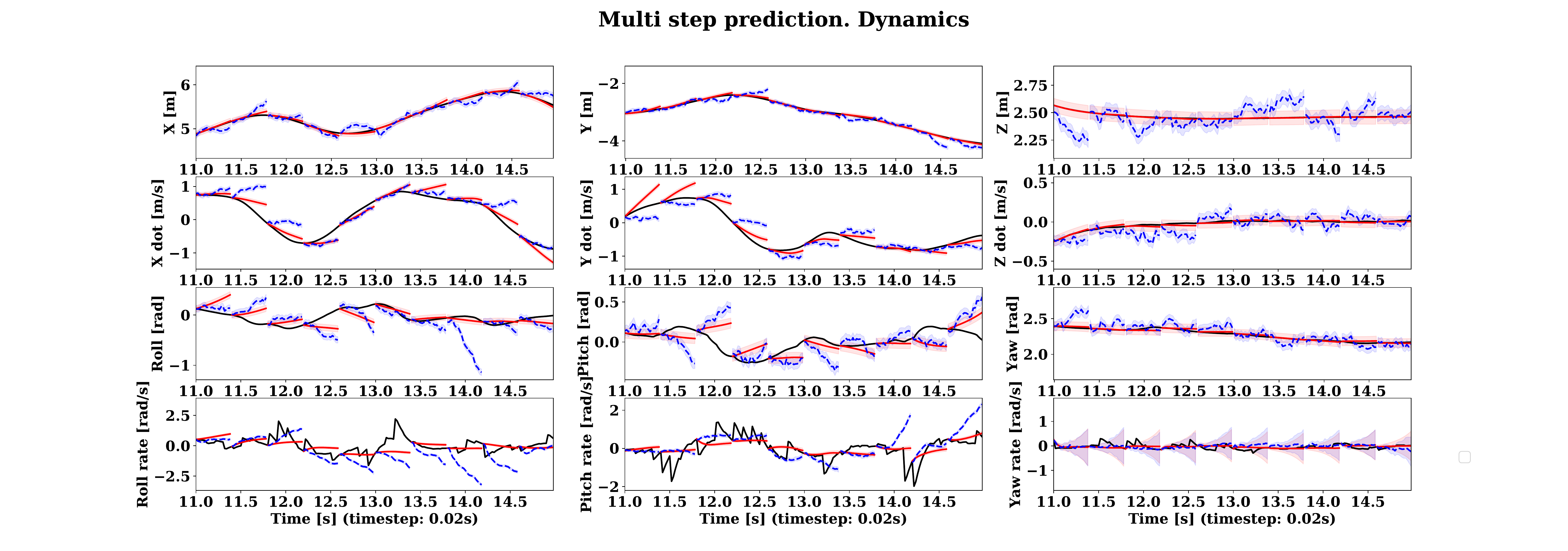} }}
  \quad
  \subfloat[Zoomed multi step prediction] {{\includegraphics[scale=0.18]{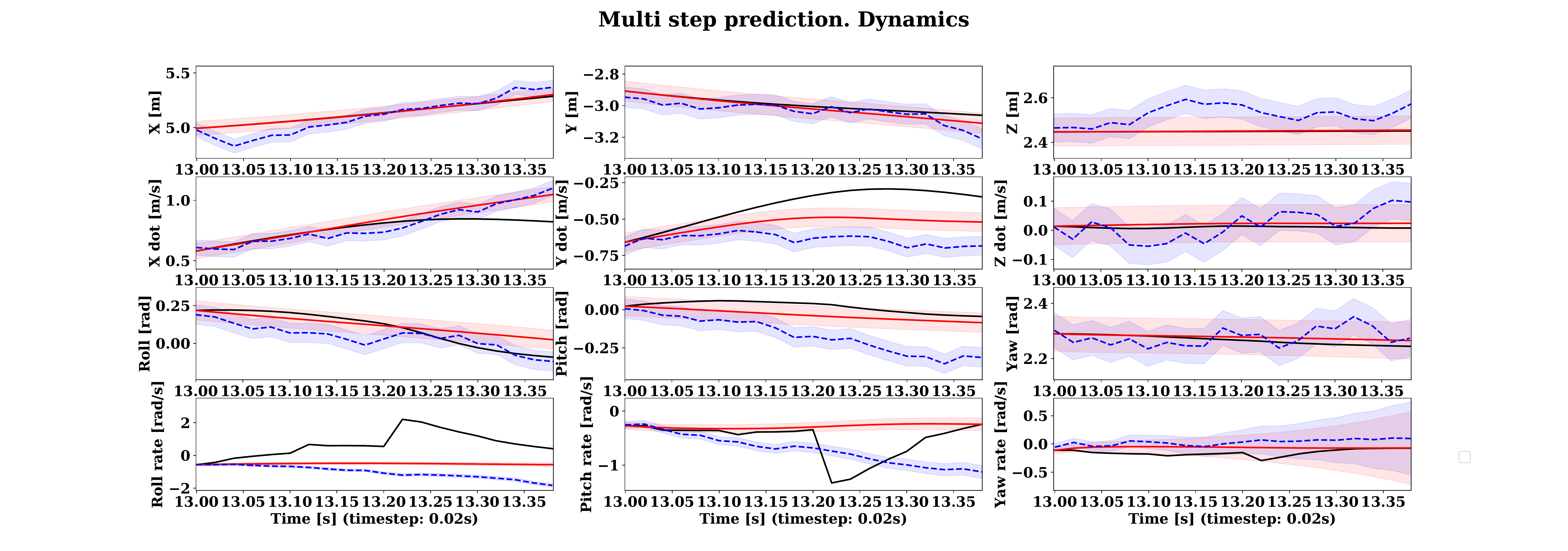} }}
  \quad
  \caption{ARDrone state transition prediction. The red line and shadow represent the predicted mean and 2$\sigma$ (standard deviation) from probabilistic ESN using spline-approximated $tanh$. The blue line and shadow are from MC sampling of deterministic ESN under uncertain inputs. Multi step prediction plots 20 time steps starting from the ground truth(black).}
  \label{fig:ARDrone}
\end{figure}

\subsubsection{Discussion}
Figure \ref{fig:CartPole} and \ref{fig:ARDrone} shows the results of single step and multi step predictions of state transitions from the learned Cart Pole and ARDrone models using probabilistic and deterministic echo state networks. As the probabilistic ESN and the deterministic ESN use the same weights of the trained model, the probabilistic ESN can propagate uncertainty successfully as Monte Carlo method does. In the single step prediction of both systems, the probabilistic ESN almost perfectly captures the output mean and variance of the MC-sampled deterministic ESN with uncertain inputs. However, in the multi step prediction, the predicted mean of probabilistic ESN diverges from the MC-sampled ESN after 4-5 time steps, but the variance of the distribution grows similarly.


\small


\end{document}